\documentclass{article}

\usepackage[table]{xcolor} 
\definecolor{cvprblue}{rgb}{0.21,0.49,0.74}
\usepackage[pagebackref,breaklinks,colorlinks,allcolors=cvprblue]{hyperref}
\PassOptionsToPackage{numbers, compress}{natbib}

\usepackage[preprint]{style}

\usepackage[utf8]{inputenc} 
\usepackage[T1]{fontenc}    
\usepackage{hyperref}       
\usepackage{url}            
\usepackage{booktabs}       
\usepackage{amsfonts}       
\usepackage{nicefrac}       
\usepackage{microtype}      
\usepackage{xcolor}         

\usepackage{graphicx}
\usepackage{booktabs}

\usepackage{amsmath}
\usepackage{cleveref}
\usepackage{multirow}
\usepackage{wrapfig}
\newcommand{\ourdataset}{Contrast-X}

\title{
Contrast-X: A Multi-Modal Contrast Image Synthesis Benchmark and   Universal Modality Flow Matching
}

\author{
Yifan Chen$^{1,*}$ 
\hspace{6mm} 
Fei Yin$^{1,*}$ \hspace{6mm} Hao Chen$^{1,*}$ \hspace{6mm} Jia Wu$^{2}$ \hspace{6mm}  Chao Li$^{1,3}$ \\
$^{1}$University of Cambridge\hspace{3mm}  $^{2}$MD Anderson Cancer Center \hspace{3mm} $^{3}$University of Dundee \\
\texttt{yifanchencam@gmail.com}
}

\newcommand{\ourbench}[0]{Contrast-X}
\newcommand{\ourmodel}[0]{FlowMI}

\usepackage{algorithm}
\usepackage{algpseudocode}
\usepackage{bbm}

\usepackage{booktabs} 
\usepackage{multirow}
\usepackage{amsthm}

\usepackage{enumitem}

\usepackage{caption}

\definecolor{lightblue}{RGB}{235,245,255} 
\definecolor{headerpurple}{RGB}{230,238,250}

\usepackage{makecell} 

\usepackage{booktabs}
\usepackage{multirow}
\usepackage{makecell}
\usepackage{threeparttable}
\usepackage{siunitx}
\newcommand{\best}[1]{\textbf{{#1}}}
\newcommand{\second}[1]{\underline{{#1}}}

\usepackage{pifont}    

\usepackage[table]{xcolor}
\usepackage{booktabs}
\usepackage{multirow}
\definecolor{ourrow}{HTML}{EDF2F8}   

\begin{document}

\maketitle

\let\thefootnote\relax\footnotetext{$^*$Yifan Chen, Fei Yin and Hao Chen contribute equally.} 

\begin{abstract}

Contrast-enhanced imaging is central to oncologic diagnosis, but contrast agents can be contraindicated for many of the patients who need them most. Synthesizing contrast scans from non-contrast inputs is the natural response. Two obstacles stand in the way: no benchmark provides paired contrast data with lesion-level evaluation, and no single model handles the arbitrary missing patterns seen in practice.  
We introduce \ourbench{}, a benchmark of paired contrast-enhanced and non-contrast imaging spanning 10 organs in CT (1{,}526 patients) and multi-phase breast DCE-MRI (1{,}116 patients). Every case carries radiologist-verified phase labels and tumor masks. 
We further propose \ourmodel{}, a single model that handles arbitrary subsets of available modalities through a unified multi-modal latent space and flow matching. 
We benchmark a range of missing-modality configurations, reporting standard image-quality metrics, radiologist reader studies, and downstream lesion analysis on the synthesized scans. We further evaluate cross-organ generalization to test whether the model has learned a transferable contrast-enhancement operation. Dataset, code, and leaderboard will be released. 
Our code are
available at https://github.com/YifanChen02/Contrast-X.
\end{abstract}




\section{Introduction}

\begin{figure}[t]
\centering
\includegraphics[width=1.0\linewidth]{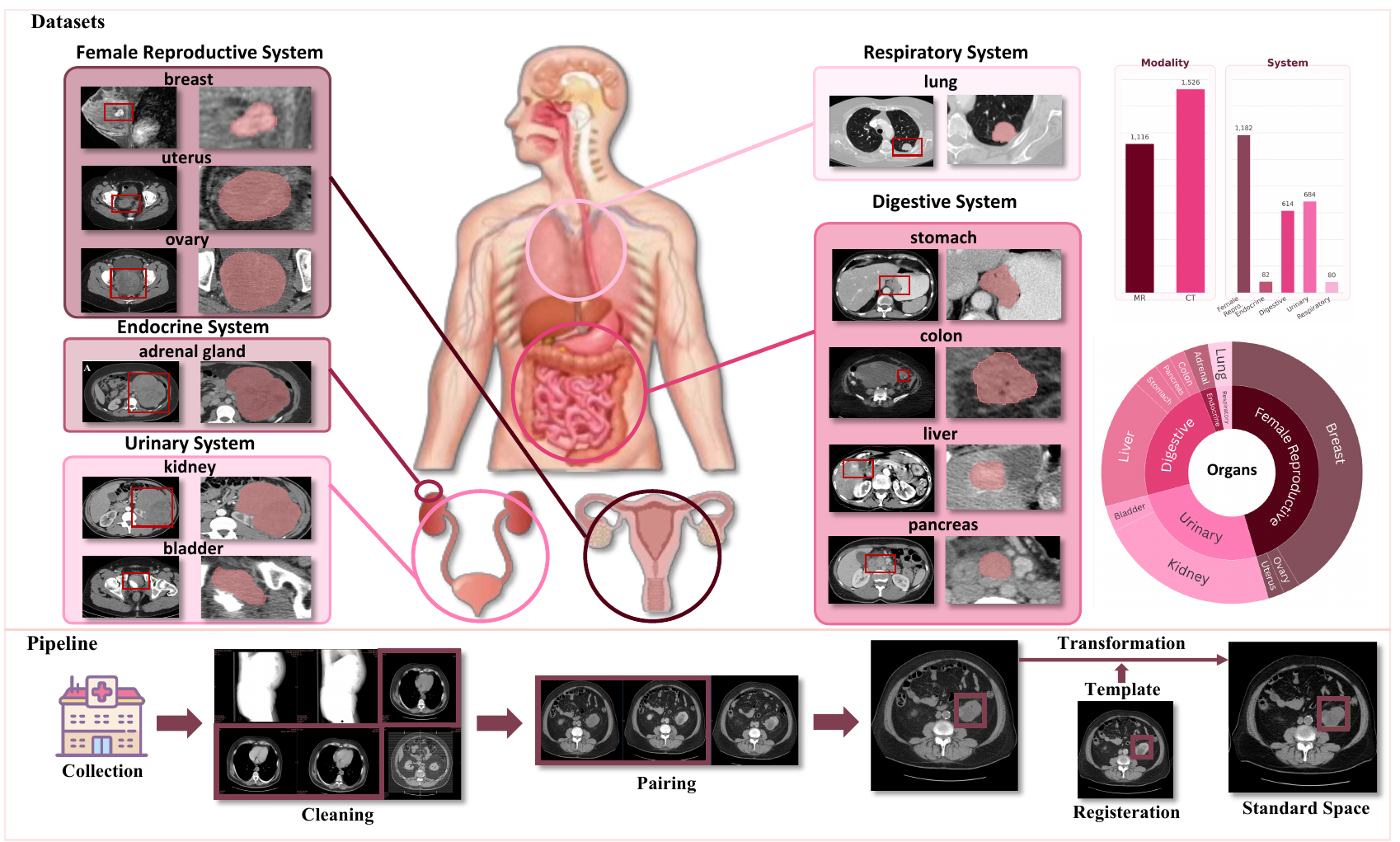}
\caption{\textit{Top:} Representative organ systems with paired
contrast and non-contrast scans. The charts show organ-wise
composition and modality balance.
\textit{Bottom:} Standardized curation pipeline including data
collection, cleaning, pairing, and registration.
}
\vspace{-2pt}
\label{fig:teaser}
\end{figure}

A radiology report is, in many cases, only as informative as the contrast 
agent that made it possible. Iodinated and gadolinium-based agents render 
tumors, vessels, and inflammation visible against tissue that would otherwise  appear uniform. They are, by quiet  consensus, the difference between a scan that diagnoses and a scan that merely shows.

Yet contrast is not a universally applicable resource. Patients with renal 
impairment, prior allergic reaction, hyperthyroidism, or pregnancy cannot 
safely receive it; many sites cannot reliably administer it; and even when 
both patient and site permit, scheduling, cost, and workflow constraints 
routinely produce studies in which the contrast portion is missing. The 
clinical consequence is a recurring asymmetry: some patients are imaged 
with the full diagnostic signal (with contrast), others without it.

Closing this asymmetry computationally has become an active research direction. If a non-contrast scan can be transformed into its contrast 
counterpart, the patient who cannot receive the agent is not penalized for 
that fact. Recent advances in image-to-image translation~\cite{I2I,Chartsias2018}, GANs~\cite{CycleGAN,Pix2Pix}, diffusion modeling~\cite{Palette,xu2024diffusion,LatentDiff,DiT}, and flow matching~\cite{lipman2023flow,chen2025learning}   produce synthetic 
medical images of plausible visual quality.

However, current contrast imaging tasks lack a benchmark, and contrast/non-contrast data (referred to as contrast pairs hereafter) is scarce and limited in scope. Many resources, such as AMOS~\cite{ji2022amos_dataset} and TCIA~\cite{TCIA}, either lack pairing or have no contrast labels (i.e., it is unclear which images are contrast-enhanced and which are not). Furthermore, they rarely include associated lesion segmentation masks for further analysis and verification of the generated results.


A second limitation is methodological. Clinical scans frequently present with missing components in arbitrary configurations: sometimes only a single phase is available, sometimes two of three. The prevailing approach is to train a dedicated model for each configuration, requiring a separate network for every source–target combination. This scales poorly (on the order of $2^N$ models for $N$ phases),  making it impractical to maintain or deploy in real-world scenarios.

In this paper, we address these two limitations that hinder the development of multi-modal contrast image synthesis (CIS). We aim to build a unified benchmark for comprehensively evaluating translation-based generative models, along with a universal method (one model for all configurations) for these tasks. In summary, our contributions are as follows: 

\begin{itemize}[leftmargin=*]
    \item We construct \ourbench, covering pan-cancer contrast pairs across 11 organs from 2{,}642 patients. Every case includes tumor masks manually delineated by radiologists and subsequently verified by oncologists. It comprises two sets of challenges: the first is a multi-organ generation task using non-contrast and contrast-enhanced computed tomography (CT/CTC) across 10 organs; the second uses multi-phase dynamic contrast-enhanced magnetic resonance imaging (DCE-MRI) sequences to evaluate different missing-modality settings. Additionally, we validate the effectiveness of the generated images through downstream segmentation tasks.
    \item  We introduce Flow Modal Imputation (FlowMI) to recast multi-modal contrast synthesis as a single transport problem in a shared latent space. Arbitrary subsets of available phases are encoded as distinct starting points, all flowing toward a common endpoint: the complete latent representation in which no modality is missing. Intermediate states along this trajectory are further constrained to semantically aligned with flow matching. 
\end{itemize}

\section{\ourdataset}
\label{sec:dataset}

\subsection{What \ourbench{} Contains}
\label{sec:dataset_overview}

\ourbench{} is assembled from public oncology imaging archives and restructured into a single benchmark with consistent format, registration, and labeling, all curated directed by senior radiologists. It contains 2{,}642 patients across 11 organs.

The benchmark is organized into two cohorts that probe different axes of contrast synthesis.

The CT cohort covers 1{,}526 patients across 10 organs frequently imaged in oncology workflows: adrenal gland, ovary, uterus, stomach, pancreas, liver, colon, bladder, kidney, and lung. Each study contains a non-contrast CT and its contrast-enhanced counterpart (CTC), representing the standard pre-/post-contrast acquisition used in abdominal and thoracic CT. The challenge here is anatomical: contrast uptake patterns differ markedly across organs, and a synthesis model is asked to generalize across them. 
The MR cohort covers 1{,}116 breast cases, each with a complete three-phase dynamic contrast-enhanced sequence (DCE1, DCE2, DCE3). Rather than a single pre-/post pair, DCE samples contrast uptake at successive time points, tracing the wash-in and wash-out kinetics that distinguish malignant from benign tissue. The challenge here is multi-phase: a synthesis model must learn the relationships across phases and recover any missing phase from the configurations that remain.

Every case in \ourbench{} carries two annotations: a radiologist-verified contrast-phase label, and a voxel-level tumor segmentation mask. The phase label resolves an ambiguity that public collections frequently leave open or mislabel. The tumor mask enables evaluation criteria that operate at the level of the clinically meaningful lesion rather than the pixel grid. We return to this point in \cref{sec:dataset_what_enables}.

\subsection{How It Was Built}
\label{sec:dataset_construction}
\ourbench{} is constructed from publicly redistributable sources. The primary collections and cohorts are summarized in Table~\ref{tb:datasets_construction}, with their licenses referenced.

\begin{table*}[t]
\centering
\caption{\textbf{Composition of \ourbench{}.} The benchmark aggregates 25 public source collections into $2{,}642$ paired cases spanning 11 organs and two domains (MR: 1{,}116 breast; CT: 1{,}526 across 10 organs). Sources are grouped by organ and listed in descending order of case count within each organ; organs themselves are ordered by total case count. All released under CC~BY~3.0, except those marked $\dagger$ (CC~BY~4.0). \#Case denotes per-source case count.}
\label{tb:datasets_construction}
\setlength{\tabcolsep}{8pt}
\renewcommand{\arraystretch}{1.25}
\resizebox{0.95\linewidth}{!}{%
\begin{tabular}{llr @{\hspace{20pt}} llr @{\hspace{20pt}} llr}
\toprule
\textbf{Source Dataset} & \textbf{Organ} & \textbf{\#Case}
& \textbf{Source Dataset} & \textbf{Organ} & \textbf{\#Case}
& \textbf{Source Dataset} & \textbf{Organ} & \textbf{\#Case} \\
\midrule
\multicolumn{9}{c}{\cellcolor{blue!5}\textit{\textbf{MR Track} — Breast, 1{,}116 cases}} \\
\midrule
I-SPY 1   & Breast & 558
& TCGA-BRCA & Breast & 378
& UCSF      & Breast & 180 \\
\midrule
\multicolumn{9}{c}{\cellcolor{blue!5}\textit{\textbf{CT Track} — 10 organs, 1{,}526 cases}} \\
\midrule
TCGA-KIRC      & Kidney  & 278
& TCGA-STAD                       & Stomach & 86
& CPTAC-PDA  & Pancreas & 54 \\
C4KC-KiTS      & Kidney  & 216
& Adrenal-ACC-Ki67-Seg$^\dagger$  & Adrenal & 82
& CPTAC-UCEC & Uterus   & 40 \\
CPTAC-CCRCC    & Kidney  & 66
& CMB-LCA$^\dagger$               & Lung    & 28
& TCGA-UCEC  & Uterus   & 14 \\
TCGA-KIRP      & Kidney  & 26
& CPTAC-LSCC                      & Lung    & 28
& CMB-CRC    & Colon    & 18 \\
TCGA-KICH      & Kidney  & 12
& CPTAC-LUAD                      & Lung    & 10
& TCGA-COAD  & Colon    & 4  \\
HCC-TACE-Seg   & Liver   & 360
& Lung-PET-CT-Dx$^\dagger$        & Lung    & 6
& TCGA-OV    & Ovary    & 12 \\
TCGA-LIHC      & Liver   & 72
& Anti-PD-1\_Lung                 & Lung    & 6
&            &          &    \\
TCGA-BLCA      & Bladder & 86
& TCGA-LUSC                       & Lung    & 2
&            &          &    \\
\bottomrule
\end{tabular}%
}
\vspace{-5pt}
\end{table*}

\textbf{Pairing.} To identify pairs, all raw image series were parsed to inspect the three metadata fields most indicative of contrast state: \texttt{SeriesDescription}, \texttt{ContrastBolusAgent}, and \texttt{AcquisitionTime}. As an initial coarse filter, studies lacking a matchable contrast pair were excluded.

\textbf{Labeling.} Each remaining study then underwent fine-grained labeling and inspection by a team of trained annotators, with final confirmation by a senior radiologist for CT versus CTC and for DCE phase identity (DCE1, DCE2, DCE3). Cases with unresolved ambiguity, non-standard or distorted imaging, severe motion artifacts, or insufficient anatomical coverage were discarded.  Tumor segmentation masks were inherited from the source dataset when available with sufficient quality and consistent with our protocol; otherwise, they were annotated by a radiologist.

\textbf{Registration.} Within each study, scan pairs were registered using the Elastix~\cite{Klein2010Elastix} ITK backend. A four-level rigid and affine pyramid was applied to all studies, followed by a three-level B-spline deformable stage for organs prone to non-rigid motion (liver, lung), with mutual information as the similarity metric throughout. Each registration was visually inspected for anatomical consistency, and cases with residual misalignment were discarded.

\textbf{Normalization.} Finally, scan volumes were resampled to isotropic $1\,\text{mm}^3$ spacing and cropped to organ-specific bounding boxes to remove irrelevant anatomy. CT intensities were windowed using organ-specific Hounsfield Unit ranges (e.g., $[-200, 300]$~HU for soft tissue) and then min-max normalized. MR intensities were standardized by per-scan z-score normalization to mitigate the inter-scanner variation typical of multi-site MR cohorts.

\subsection{What Makes It Different}
\label{sec:dataset_comparison}

\begin{table}[]
\setlength{\tabcolsep}{4pt}
\caption{Comparison of \protect\ourdataset{} with existing medical imaging benchmarks. 
``CE Pair'' denotes explicit per-patient pairing of contrast-enhanced and non-contrast scans; 
``Multi-Phase'' indicates availability of multiple temporal contrast phases (e.g., DCE); 
``Mask'' denotes availability of voxel-level segmentation annotations (anatomical structures or lesions); 
``\#Organs'' counts distinct organs covered;
``Size'' reports the number of patients.
$\bullet$ = full support, $\circ$ = partial support, --- = not provided. $\sim$ is the estimated number. }
\label{tb:comparison-benchmarks}
\small
\centering
\resizebox{1.0\textwidth}{!}{
\begin{tabular}{l|cc|ccc|cc|l}
\toprule
\textbf{Benchmark} 
& \textbf{MR} 
& \textbf{CT} 
& \textbf{CE Pair} 
& \textbf{Multi-Phase} 
& \textbf{Mask}
& \textbf{\#Organs} 
& \textbf{Size} 
& \textbf{Primary Application} \\
\midrule
\multicolumn{9}{l}{\textit{Brain-centric benchmarks}} \\
\midrule
BraTS-GLI 2024 \cite{BraTS2024} & $\bullet$ & ---       & $\bullet$ & ---       & $\bullet$ & 1 & 1{,}350 & Post-treatment glioma seg. \\
BraSyn 2023 \cite{BraSyn}           & $\bullet$ & ---       & $\circ$   & ---       & $\bullet$ & 1 & 1{,}251 & Missing-sequence synthesis \\
IXI \cite{IXI}                      & $\bullet$ & ---       & ---       & ---       & ---       & 1 & $\sim$600  & Healthy brain reference \\
OASIS-3 \cite{OASIS-3}              & $\bullet$ & ---       & ---       & ---       & $\circ$   & 1 & 1{,}378 & Alzheimer's research \\
ADNI-3 \cite{ADNI-3}                & $\bullet$ & ---       & ---       & ---       & $\circ$   & 1 & $\sim$1{,}500  & Alzheimer's research \\
AANLIB \cite{AANLIB}                & $\bullet$ & $\bullet$ & ---       & ---       & ---       & 1 & $\sim$30 & Normal \& major brain diseases \\
\midrule
\multicolumn{9}{l}{\textit{Region-specific / cross-modality benchmarks}} \\
\midrule
CHAOS \cite{CHAOS}                  & $\bullet$ & $\bullet$ & ---       & ---       & $\bullet$ & 4 & 80    & Healthy abdominal organ seg. \\
crossMoDA \cite{CrossMoDA}          & $\bullet$ & ---       & ---       & ---       & $\bullet$ & 1 & 379   & Cochlear domain adaptation \\
ACDC \cite{ACDC}                    & $\bullet$ & ---       & ---       & ---       & $\bullet$ & 1 & 150   & Cardiac diagnosis \\
MMWHS \cite{MMWHS}                  & $\bullet$ & $\bullet$ & ---       & ---       & $\bullet$ & 1 & 120   & Cardiac structure seg. \\
SynthRAD2023 \cite{SynthRAD2023}    & $\bullet$ & $\bullet$ & $\bullet$ & ---       & $\circ$   & 2 & 1{,}080 & MR-to-CT synthesis \\
\midrule
\multicolumn{9}{l}{\textit{Whole-body / PET benchmarks}} \\
\midrule
FDG-PET/CT \cite{FDG-PET/CT}        & ---       & $\bullet$ & ---       & ---       & $\bullet$ & --- & 900    & PET--CT tumor lesion analysis \\
Ultra-low Dose \cite{UDPET}         & ---       & ---       & ---       & ---       & ---       & --- & 1{,}447 & Low-dose PET synthesis \\
\midrule
\multicolumn{9}{l}{\textit{Multi-organ contrast-pair benchmarks}} \\
\midrule
\rowcolor{gray!15}
\textbf{\ourdataset{} (Ours)}       & $\bullet$ & $\bullet$ & $\bullet$ & $\bullet$ & $\bullet$ & \textbf{11} & \textbf{2{,}642} & \textbf{Pan-cancer contrast synthesis} \\
\bottomrule
\end{tabular}
}
\vspace{-10pt}
\end{table}

Medical image translation has progressed largely on benchmarks designed for adjacent problems, and none provides the per-patient contrast pairing across organs and modalities that clinical oncology imaging requires. Brain MR datasets dominate the field: BraTS~\cite{BraTS2024} offers T1/T1ce pairs that have powered contrast-synthesis research, and IXI~\cite{IXI}, OASIS-3~\cite{OASIS-3}, and ADNI~\cite{ADNI-3} have driven cross-sequence MR synthesis. Yet all are confined to the brain, and methods developed on them do not transfer to multi-organ settings. Beyond the brain, whole-body resources such as FDG-PET/CT~\cite{FDG-PET/CT} and UDPET~\cite{UDPET} achieve impressive scale, but they pair PET with CT rather than contrast pair, addressing a fundamentally different translation task. Cross-modality benchmarks like MMWHS~\cite{MMWHS} provide MR and CT data, but do not provide the supervised contrast pairing our setting requires.

\ourbench{} fills this gap by combining the three properties no prior benchmark offers together: explicit per-patient pairing of contrast and non-contrast scans, broad anatomical coverage across both CT and MR, and tumor-level annotation for clinically meaningful evaluation. Table~\ref{tb:comparison-benchmarks} summarizes the comparison along five axes.

\textbf{\textit{Note.}} The term ``multi-modal'' carries different meanings across the literature, referring variously to distinct imaging modalities (e.g., MR and CT), multiple MR sequences within a study (as in BraTS), or multi-phase CT acquisitions. For clarity, throughout this paper we use \emph{modality} to denote an individual MR sequence or CT phase, and \emph{multi-domain} to denote the MR--CT setting.

 \subsection{What It Benchmarks}
\label{sec:dataset_what_enables}

\begin{figure*}[t]
\begin{center}
\centerline{\includegraphics[width=1\linewidth]{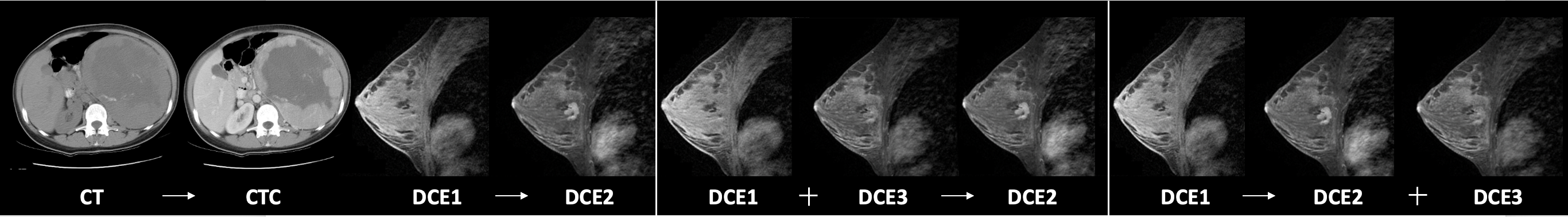}}
\caption{Representative tasks instantiated on \ourbench{}. 
(a) CT $\to$ CTC: cross-organ contrast synthesis on the CT cohort. 
(b) DCE$_1 \to$ DCE$_2$: neighbor-phase prediction on the MR cohort. 
(c) DCE$_{1,3} \to$ DCE$_2$: recovery of an intermediate phase from its temporal neighbors. 
(d) DCE$_1 \to$ DCE$_{2,3}$: synthesis of later phases from the earliest acquired phase. 
Settings (a) and (b) are \emph{one-to-one}; (c) and (d) instantiate the more general case in which an arbitrary subset of modalities is observed and a disjoint subset is predicted (Table~\ref{sec:dataset_what_enables}).}

\label{fig:task_definitions}
\vspace{-11pt}
\end{center}
\end{figure*}

\ourbench{} defines a family of contrast-synthesis tasks, which the two cohorts instantiate along complementary axes (Figure~\ref{fig:task_definitions}). 

\noindent\textbf{CT cohort: cross-organ contrast synthesis.} The CT cohort defines a one-to-one task, CT $\to$ CTC, in which a non-contrast scan predicts its contrast-enhanced counterpart. It tests whether a single model can learn a contrast mapping that generalizes across anatomy rather than memorizing organ-specific behavior. We evaluate under two protocols: a full-organ setting that trains and tests on all organs jointly, and a leave-one-organ-out setting that probes generalization to unseen anatomy.

\noindent\textbf{MR cohort: multi-phase DCE synthesis.} The MR cohort captures the temporal dynamics of contrast enhancement through three tasks over the phases DCE$_1$, DCE$_2$, DCE$_3$ (collectively denoted DCE$_{1,2,3}$). DCE$_1 \to$ DCE$_2$ tests direct one-to-one prediction, mirroring the CT $\to$ CTC setting. DCE$_{1,3} \to$ DCE$_2$ and DCE$_1 \to$ DCE$_{2,3}$ reflect the different clinical scenarios in which only a subset of phases is acquired and the remainder must be synthesized. We exploit this view in Section~\ref{sec:method} to motivate a single model trained jointly across all configurations.

\noindent\textbf{Splits.} 
\label{sec:dataset_splits}
We partition \ourbench{} at the organ-level into training (70\%), validation (10\%), and test (20\%) sets, using stratified sampling to preserve the distribution over organ systems and cancer types.  Main results in Section~\ref{exp:experiments} and ablations are reported on the full test set. 



\begin{figure*}[t]
\begin{center}
\centerline{\includegraphics[width=1\linewidth]{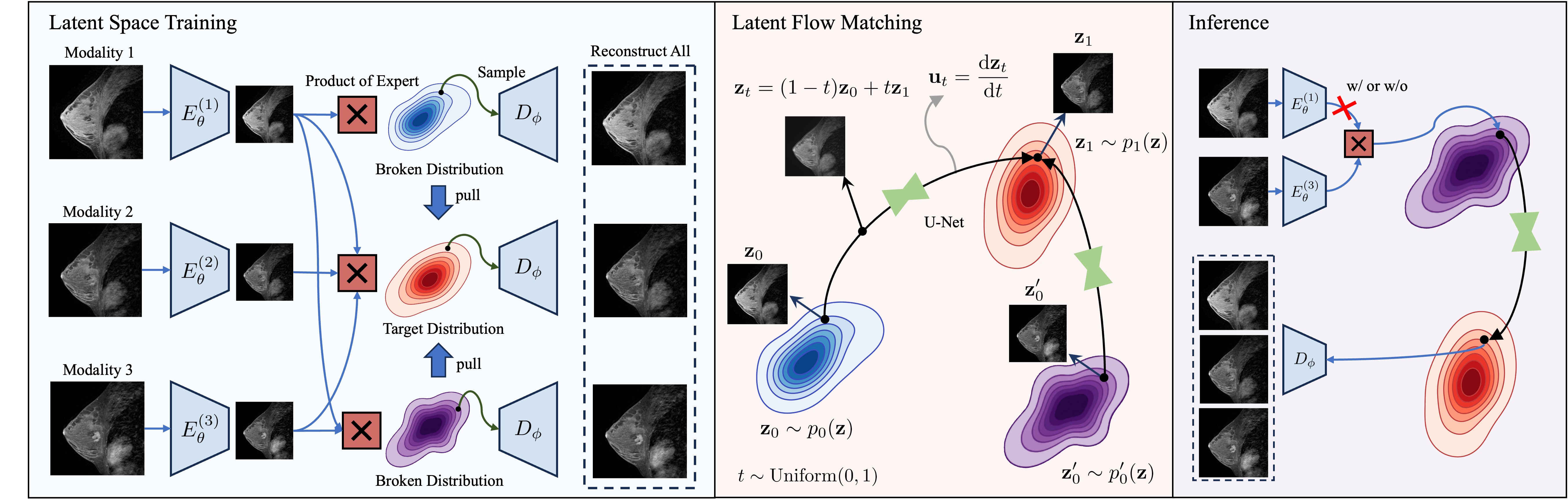}}
\caption{Overview of the proposed FlowMI framework. 
\textbf{Left:} Modality-specific encoders $E_\theta^{(i)}$ map inputs into a latent space, which are fused via a product-of-experts. The distribution with all modalities defines the \emph{target}, while cases with missing modalities define the \emph{broken} distribution. 
\textbf{Middle:} Latent flow matching learns a smooth mapping from $p_0(z)$ (broken) to $p_1(z)$ (target) using a U-Net parameterization of the velocity field, $\mathbf{u}_t = \tfrac{dz_t}{dt}$. 
\textbf{Right:} During inference, inputs with missing modalities are encoded and aligned through the learned flow, enabling consistent reconstruction or synthesis of complete modalities.}
\label{fig:architecture}
\vspace{-12pt}
\end{center}
\end{figure*}

\section{Flow Modality Imputation (FlowMI) }
\label{sec:method}

We propose \textbf{FlowMI}, which unifies autoencoding and flow matching into a single framework for missing-modality imputation. The core idea is to treat any partial-modality configuration as a starting latent and the full-modality observation as the endpoint, with flow matching transporting one to the other. We further shape the latent space so that intermediate states along the flow remain meaningful, aligning the trajectory with the geometry of the latent manifold. We want to highlight that our contribution is not the use of flow matching or Product-of-Experts themselves, but rather the reformulation of these modules and the proposed training scheme that together make FlowMI a unified model capable of handling arbitrary observation configurations.

\subsection{Preliminary: Flow Matching}
\label{sec:method_preliminary}

Flow Matching (FM)~\cite{lipman2023flow,liu2022flow} learns a continuous transport map between two distributions $\pi_0$ and $\pi_1$ over a space $\mathcal{Z} \subseteq \mathbb{R}^d$ from samples alone. A time-dependent vector field $v_\theta(\mathbf{z}, t)$ is trained to follow a prescribed interpolation path $\mathbf{z}_t$ between paired endpoints $(\mathbf{z}_0, \mathbf{z}_1) \sim \pi_0 \times \pi_1$ by regressing the analytic target velocity:
\begin{equation}
\label{eq:fm_loss_prelim}
\mathcal{L}_{\mathrm{FM}}(\theta) \;=\; \mathbb{E}_{t,\,(\mathbf{z}_0,\mathbf{z}_1)} \left[\, \big\| v_\theta(\mathbf{z}_t, t) - \dot{\mathbf{z}}_t \big\|^2 \,\right],
\end{equation}
with $t \sim \mathcal{U}(0,1)$. We adopt straight-line interpolation $\mathbf{z}_t = (1-t)\mathbf{z}_0 + t\mathbf{z}_1$, for which the target velocity reduces to the constant $\mathbf{z}_1 - \mathbf{z}_0$. Once trained, $v_\theta$ transports samples by integrating $\mathrm{d}\mathbf{z}_t/\mathrm{d}t = v_\theta(\mathbf{z}_t, t)$ from $t=0$ to $1$. Latent Flow Matching~\cite{dao2023flow,chadebec2025lbm} applies this construction inside the latent space of a pretrained autoencoder. FlowMI extends this construction to a clean missing-modality imputation regime: both endpoints are produced by \emph{the same autoencoder under different observation masks}, casting any missing-modality imputation as a transport problem on a unify learned manifold.

\subsection{Unified Latent Space via Product-of-Experts Autoencoder}
\label{sec:method_ae}

A latent space that supports flow transport across observation regimes should satisfy two requirements: it can accept any subset of modalities without architectural changes, and it can provide a continuous manifold aligned with the flow trajectory. We meet both requirements with a unified latent space design.

For modality flexibility, we adopt a multi-modal VAE with modality-specific encoders and a shared decoder, and fuse the encoded distributions via a Product-of-Experts (PoE)~\cite{wu2018mvae} to yield a latent space of fixed dimensionality regardless of which modality configuration is observed.

Let a multi-modal sample $\mathbf{x} = \{x^{(1)}, \dots, x^{(M)}\}$ contains $M$ modalities, indexed by $\mathcal{M} = \{1, \dots, M\}$. A binary mask $\mathbf{m} \in \{0,1\}^M$ encodes observability, with $\mathbf{m}^{(i)} = 1$ if modality $i$ is observed and $0$ otherwise. We write $\mathcal{O}(\mathbf{m}) = \{i : \mathbf{m}^{(i)} = 1\}$ for the observed index set, $\mathbf{x}^{\mathbf{m}}$ for the observed subset, and $\mathbf{x}^{\mathbbm{1}}$ for the fully observed sample, where $\mathbbm{1}$ is the all-ones mask. The objective is to predict $\mathbf{x}^{\neg\mathbf{m}}$ from $\mathbf{x}^{\mathbf{m}}$.

For each modality $m$, an encoder $E^{(m)}_\theta$ outputs a Gaussian expert capturing what modality $m$ alone reveals about the shared latent $\mathbf{z} \in \mathbb{R}^d$:
\begin{equation}
\label{eq:expert}
    q_\theta^{(m)}(\mathbf{z}\mid x^{(m)}) = \mathcal{N}\!\big(\mathbf{z};\, \boldsymbol{\mu}^{(m)},\, \mathrm{diag}(\boldsymbol{\sigma}^{(m)})^2\big), \qquad (\boldsymbol{\mu}^{(m)}, \boldsymbol{\sigma}^{(m)}) = E_\theta^{(m)}(x^{(m)}).
\end{equation}
Given an observation mask $\mathbf{m}$, we fuse only the experts of observed modalities with the prior $p(\mathbf{z}) = \mathcal{N}(\mathbf{0}, \mathbf{I})$:
\begin{equation}
\label{eq:poe}
    q_\theta(\mathbf{z}\mid \mathbf{x}^{\mathbf{m}}) \propto p(\mathbf{z}) \prod_{m \in \mathcal{O}(\mathbf{m})} q_\theta^{(m)}(\mathbf{z}\mid x^{(m)}).
\end{equation}

The prior plays two roles: it keeps the posterior well-defined for any mask (including when few or no modalities are observed) and provides the standard VAE regularization target. Since the product of Gaussians is Gaussian, the posterior has closed form,
\begin{equation}
\label{eq:poe_closed_form}
    \boldsymbol{\Lambda}^{\mathbf{m}} = \mathbf{I} + \!\!\sum_{m \in \mathcal{O}(\mathbf{m})}\!\! \mathrm{diag}(\boldsymbol{\sigma}^{(m)})^{-2}, \qquad
    \boldsymbol{\mu}^{\mathbf{m}} = (\boldsymbol{\Lambda}^{\mathbf{m}})^{-1} \!\!\sum_{m \in \mathcal{O}(\mathbf{m})}\!\! \mathrm{diag}(\boldsymbol{\sigma}^{(m)})^{-2}\, \boldsymbol{\mu}^{(m)},
\end{equation}
so each expert contributes proportionally and the prior  anchors the posterior when evidence is sparse.

We then sample the fused code via reparameterization, $\mathbf{z}^{\mathbf{m}} = \boldsymbol{\mu}^{\mathbf{m}} + (\boldsymbol{\Lambda}^{\mathbf{m}})^{-1/2}\boldsymbol{\epsilon}$ with $\boldsymbol{\epsilon} \sim \mathcal{N}(\mathbf{0}, \mathbf{I})$ at training, or take $\mathbf{z}^{\mathbf{m}} = \boldsymbol{\mu}^{\mathbf{m}}$ at inference. A single decoder $D_\phi$ then maps $\mathbf{z}^{\mathbf{m}}$ to image space for reconstruction.

The autoencoding process ensures that the endpoints $\mathbf{z}_0$ and $\mathbf{z}_1$ are meaningful, but the flow trajectory between them is not. In practice, latent flow matching never lands exactly on $\mathbf{z}_1$: the integrated trajectory drifts, and any drift in latent space is amplified by the decoder into unpredictable image artifacts. The only way to make this drift tolerable is to ensure that every point along the path, not just the endpoints, decodes to something meaningful. We therefore require the latent trajectory to mirror a known, well-behaved trajectory in image space.

Let $\mathbf{x}_0$ denote a missing-modality observation and $\mathbf{x}_1$ its complete-modality counterpart, with corresponding latents $\mathbf{z}_0$ and $\mathbf{z}_1$ obtained from the PoE encoder. We construct the image-space interpolant $\mathbf{x}_t = (1-t)\mathbf{x}_0 + t\mathbf{x}_1$ and the latent interpolant $\mathbf{z}_t = (1-t)\mathbf{z}_0 + t\mathbf{z}_1$, and require that $D_\phi(\mathbf{z}_t)$ recovers $\mathbf{x}_t$ for every $t \in [0,1]$:
\begin{equation}
\label{eq:rec_loss}
    \mathcal{L}_{\text{Trajectory}} = \mathbb{E}_{\mathbf{x},\,\mathbf{m},\,t}\!\left[\, \big\| D_\phi(\mathbf{z}_t) - \mathbf{x}_t \big\|_2^2 \,\right].
\end{equation}
In practice, we apply a stop-gradient on $\mathbf{z}_0$ and $\mathbf{z}_1$ when forming $\mathbf{z}_t$ to stabilize training.
This per-step reconstruction constraint anchors the entire latent path to image-space ground truth, so the flow has no opportunity to drift into a region the decoder cannot interpret. Every intermediate $\mathbf{z}_t$ remains semantically meaningful and decodable, and the latent trajectory becomes a faithful counterpart to the image-space flow rather than an unconstrained curve through latent space.

\subsection{Latent Flow Modality Imputation}
\label{sec:method_flow}

The autoencoder defines, for every sample $\mathbf{x}$ and every mask $\mathbf{m}$, a paired endpoint structure $(\mathbf{z}^{\mathbf{m}}, \mathbf{z}^{\mathbbm{1}})$ on a shared latent manifold. We model imputation as a flow between these endpoints, conditioned on the mask so that a single flow field handles arbitrary missing patterns. Concretely, we define a mask-conditioned vector field $v_\theta(\mathbf{z}, t, \mathbf{m})$ and use straight-line interpolation between endpoints,
\begin{equation}
\label{eq:interp}
    \mathbf{z}_t = (1-t)\,\mathbf{z}^{\mathbf{m}} + t\,\mathbf{z}^{\mathbbm{1}},
\end{equation}
for $t \in [0,1]$, training the flow by regressing the constant target velocity:
\begin{equation}
\label{eq:lfm_loss}
    \mathcal{L}_{\text{LFM}} = \mathbb{E}_{\mathbf{x},\,\mathbf{m},\,t}\!\left[\, \big\| v_\theta(\mathbf{z}_t,\, t,\, \mathbf{m}) - (\mathbf{z}^{\mathbbm{1}} - \mathbf{z}^{\mathbf{m}}) \big\|^2 \,\right],
\end{equation}
where $\mathbf{m}$ is sampled uniformly over non-empty subsets of $\mathcal{M}$ at each step.

\paragraph{Inference.} 
\label{sec:method_inference}
Given a test input $\mathbf{x}^{\mathbf{m}}$ with mask $\mathbf{m}$, we encode the partial-observation latent $\mathbf{z}_0 = E_\theta(\mathbf{x}^{\mathbf{m}}, \mathbf{m})$ and integrate the learned ODE $\mathrm{d}\mathbf{z}_t/\mathrm{d}t = v_\theta(\mathbf{z}_t, t, \mathbf{m})$ from $t=0$ to $t=1$ using a standard Euler scheme,
\begin{equation}
    \mathbf{z}_{t+\Delta t} \;=\; \mathbf{z}_t + \Delta t \cdot v_\theta(\mathbf{z}_t, t, \mathbf{m}), \qquad t = 0, \Delta t, \dots, 1 - \Delta t.
\end{equation}
The terminal state $\hat{\mathbf{z}}^{\mathbbm{1}} := \mathbf{z}_1$ approximates the full-observation latent, which the decoder maps to a complete reconstruction $\hat{\mathbf{x}}^{\mathbbm{1}} = D_\phi(\hat{\mathbf{z}}^{\mathbbm{1}})$. The imputed missing modalities $\hat{\mathbf{x}}^{\neg\mathbf{m}}$ are read off from the corresponding output channels.

\begin{table*}[t]
\centering
\scriptsize
\setlength{\tabcolsep}{3pt}
\renewcommand{\arraystretch}{1.32}
\caption{
Quantitative comparison on the \protect\ourdataset{} CT and MR paired pan-cancer contrast media dataset. Methods are grouped by generative mechanism (Direct regression, GAN, Diffusion, Flow Matching) and backbone architecture. PSNR (dB) and SSIM (\%) are reported as Mean$\pm$Std across the test cohort. \best{Bold} marks the best result per column; \second{underline} marks the second-best. *Trans: Transformer; Arch.: Architecture.
}
\label{tb:experiments_mechanism-based}

\resizebox{\textwidth}{!}{%
\begin{tabular}{@{} l|c|c| cc | cc | cc | cc @{}}
\toprule
\multirow{2}{*}{\textbf{Mechanism}}
& \multirow{2}{*}{\textbf{Arch.}}
& \multirow{2}{*}{\textbf{Method}}
& \multicolumn{2}{c|}{\textbf{CT $\rightarrow$ CTC}}
& \multicolumn{2}{c|}{\textbf{DCE$_1$ $\rightarrow$ DCE$_2$}}
& \multicolumn{2}{c|}{\textbf{DCE$_1$ $\rightarrow$ DCE$_{2,3}$}}
& \multicolumn{2}{c}{\textbf{DCE$_{1,3}$ $\rightarrow$ DCE$_2$}} \\
\cmidrule(lr){4-5} \cmidrule(lr){6-7} \cmidrule(lr){8-9} \cmidrule(l){10-11}
& &
& \textbf{PSNR}$\uparrow$ & \textbf{SSIM(\%)}$\uparrow$
& \textbf{PSNR}$\uparrow$ & \textbf{SSIM(\%)}$\uparrow$
& \textbf{PSNR}$\uparrow$ & \textbf{SSIM(\%)}$\uparrow$
& \textbf{PSNR}$\uparrow$ & \textbf{SSIM(\%)}$\uparrow$ \\
\midrule

\multirow{3}{*}{Direct}
& UNet  & HiNet~\cite{HiNet}
& 22.23\,\tiny{$\pm$3.70} & 77.7\,\tiny{$\pm$10.3}
& 23.41\,\tiny{$\pm$3.48} & 68.1\,\tiny{$\pm$11.9}
& 23.88\,\tiny{$\pm$3.37} & 72.9\,\tiny{$\pm$8.32}
& 26.47\,\tiny{$\pm$3.32} & 70.2\,\tiny{$\pm$12.2} \\

& Trans & ResViT~\cite{ResViT}
& 20.80\,\tiny{$\pm$3.11} & 74.2\,\tiny{$\pm$10.2}
& 25.12\,\tiny{$\pm$3.16} & 70.5\,\tiny{$\pm$10.5}
& 24.65\,\tiny{$\pm$2.81} & 63.5\,\tiny{$\pm$10.8}
& 25.24\,\tiny{$\pm$2.89} & 59.8\,\tiny{$\pm$15.0} \\

& Mamba & I2IMamba~\cite{I2I}
& 20.97\,\tiny{$\pm$3.20} & 74.6\,\tiny{$\pm$10.3}
& 23.25\,\tiny{$\pm$2.37} & 55.9\,\tiny{$\pm$14.5}
& 23.57\,\tiny{$\pm$2.50} & 58.0\,\tiny{$\pm$13.0}
& 26.82\,\tiny{$\pm$3.32} & 70.1\,\tiny{$\pm$11.8} \\
\midrule

\multirow{2}{*}{GAN}
& UNet & CycleGAN~\cite{CycleGAN}
& 21.90\,\tiny{$\pm$3.98} & 75.8\,\tiny{$\pm$13.2}
& 24.18\,\tiny{$\pm$3.33} & 65.7\,\tiny{$\pm$13.3}
& 24.46\,\tiny{$\pm$3.52} & 71.4\,\tiny{$\pm$12.0}
& 25.74\,\tiny{$\pm$3.28} & 72.4\,\tiny{$\pm$12.2} \\

& UNet & Pix2Pix~\cite{Pix2Pix}
& 21.39\,\tiny{$\pm$3.10} & 72.6\,\tiny{$\pm$13.8}
& 23.24\,\tiny{$\pm$3.35} & 64.6\,\tiny{$\pm$14.3}
& 23.90\,\tiny{$\pm$2.97} & 72.2\,\tiny{$\pm$8.95}
& 26.42\,\tiny{$\pm$3.10} & 70.2\,\tiny{$\pm$10.4} \\
\midrule

\multirow{2}{*}{Diffusion}
& UNet & Palette~\cite{Palette}
& 15.88\,\tiny{$\pm$5.30} & 33.2\,\tiny{$\pm$21.8}
& 9.54\,\tiny{$\pm$2.17}  & 9.70\,\tiny{$\pm$6.38}
& 10.19\,\tiny{$\pm$1.99} & 14.5\,\tiny{$\pm$6.16}
& 4.79\,\tiny{$\pm$1.25}  & 5.47\,\tiny{$\pm$2.55} \\

& UNet & SelfRDB~\cite{selfRDB}
& 17.16\,\tiny{$\pm$2.38} & 72.5\,\tiny{$\pm$7.90}
& 23.40\,\tiny{$\pm$3.31} & 57.6\,\tiny{$\pm$16.4}
& 24.98\,\tiny{$\pm$2.87} & 67.6\,\tiny{$\pm$13.5}
& 26.66\,\tiny{$\pm$3.79} & 70.2\,\tiny{$\pm$11.9} \\
\midrule

\multirow{4}{*}{Flow Matching}
& UNet & ConcatFM~\cite{lipman2023flow}
& \second{23.10\,\tiny{$\pm$4.12}} & \second{77.1\,\tiny{$\pm$9.28}}
& \second{26.31\,\tiny{$\pm$2.67}} & \second{72.9\,\tiny{$\pm$6.61}}
& \second{26.25\,\tiny{$\pm$2.49}} & \second{71.8\,\tiny{$\pm$6.18}}
& \second{29.06\,\tiny{$\pm$2.73}} & \second{76.1\,\tiny{$\pm$6.63}} \\

& UNet & DirectFM~\cite{lipman2023flow}
& 22.84\,\tiny{$\pm$3.88} & 76.7\,\tiny{$\pm$11.4}
& 25.74\,\tiny{$\pm$3.14} & 68.5\,\tiny{$\pm$8.09}
& 25.98\,\tiny{$\pm$3.05} & 71.1\,\tiny{$\pm$8.22}
& 27.90\,\tiny{$\pm$3.18} & 74.2\,\tiny{$\pm$7.69} \\

& UNet & PMRF~\cite{PMRF}
& 21.91\,\tiny{$\pm$3.95} & 76.6\,\tiny{$\pm$11.6}
& 25.06\,\tiny{$\pm$3.60} & 67.5\,\tiny{$\pm$8.28}
& 26.11\,\tiny{$\pm$3.80} & 70.1\,\tiny{$\pm$8.32}
& 27.53\,\tiny{$\pm$4.25} & 74.9\,\tiny{$\pm$5.31} \\

\rowcolor{ourrow}
& UNet & \textbf{FlowMI (Ours)}
& \best{24.47\,\tiny{$\pm$4.15}} & \best{78.5\,\tiny{$\pm$8.62}}
& \best{26.52\,\tiny{$\pm$3.13}} & \best{74.2\,\tiny{$\pm$9.33}}
& \best{26.63\,\tiny{$\pm$3.11}} & \best{73.7\,\tiny{$\pm$8.06}}
& \best{29.17\,\tiny{$\pm$3.24}} & \best{76.2\,\tiny{$\pm$7.65}} \\
\bottomrule
\end{tabular}
}
\vspace{-5pt}
\end{table*}

\section{Experiments}
\label{exp:experiments}

We benchmark four categories of generative models on \ourdataset{}, covering both multimodal translation and missing-modality synthesis:
{(i) Direct regression}: HiNet~\cite{HiNet}, ResViT~\cite{ResViT}, and I2IMamba~\cite{I2I};
{(ii) GAN-based}: CycleGAN~\cite{CycleGAN} and Pix2Pix~\cite{Pix2Pix};
{(iii) Diffusion-based}: Palette~\cite{Palette} and SelfRDB~\cite{selfRDB};
{(iv) Flow matching}: ConcatFM~\cite{lipman2023flow}, DirectFM~\cite{lipman2023flow}, PMRF~\cite{PMRF}, and our {FlowMI}.
We use each method's public implementation throughout.


\begin{table*}[t]
\centering
\scriptsize
\setlength{\tabcolsep}{3pt}
\renewcommand{\arraystretch}{1.32}
\caption{
LPIPS, FID, and KID comparison across methods on \protect\ourdataset{}. Methods are grouped by generative mechanism and backbone architecture. \best{Bold} marks the best result per column; \second{underline} marks the second-best. *Trans: Transformer; Arch.: Architecture.
}
\label{tb:experiments_lpips_fid_kid}

\resizebox{\textwidth}{!}{%
\begin{tabular}{@{} l|c|c| ccc | ccc | ccc | ccc @{}}
\toprule
\multirow{2}{*}{\textbf{Mechanism}}
& \multirow{2}{*}{\textbf{Arch.}}
& \multirow{2}{*}{\textbf{Method}}
& \multicolumn{3}{c|}{\textbf{CT $\rightarrow$ CTC}}
& \multicolumn{3}{c|}{\textbf{DCE$_1$ $\rightarrow$ DCE$_2$}}
& \multicolumn{3}{c|}{\textbf{DCE$_1$ $\rightarrow$ DCE$_{2,3}$}}
& \multicolumn{3}{c}{\textbf{DCE$_{1,3}$ $\rightarrow$ DCE$_2$}} \\
\cmidrule(lr){4-6} \cmidrule(lr){7-9} \cmidrule(lr){10-12} \cmidrule(l){13-15}
& &
& LPIPS$\downarrow$ & FID$\downarrow$ & KID$\downarrow$
& LPIPS$\downarrow$ & FID$\downarrow$ & KID$\downarrow$
& LPIPS$\downarrow$ & FID$\downarrow$ & KID$\downarrow$
& LPIPS$\downarrow$ & FID$\downarrow$ & KID$\downarrow$ \\
\midrule

\multirow{3}{*}{Direct}
& UNet  & HiNet~\cite{HiNet}
& 0.084 & 0.01 & 14.62
& 0.181 & 0.02 & 37.68
& 0.163 & 0.01 & 30.63
& 0.113 & 0.01 & 18.83 \\

& Trans & ResViT~\cite{ResViT}
& 0.109 & 0.01 & 17.53
& 0.224 & 0.02 & 43.18
& 0.146 & 0.01 & \second{18.31}
& 0.142 & 0.01 & 31.08 \\

& Mamba & I2IMamba~\cite{I2I}
& 0.099 & \best{0.00} & \second{12.21}
& 0.217 & 0.02 & 40.17
& 0.181 & 0.01 & 28.88
& \second{0.099} & \best{0.00} & \best{12.21} \\
\midrule

\multirow{2}{*}{GAN}
& UNet & CycleGAN~\cite{CycleGAN}
& \second{0.079} & \best{0.00} & \best{8.98}
& \second{0.150} & 0.01 & 20.88
& 0.159 & 0.01 & 22.25
& 0.114 & 0.01 & 21.40 \\

& UNet & Pix2Pix~\cite{Pix2Pix}
& 0.109 & 0.01 & 22.70
& 0.218 & 0.02 & 48.92
& 0.186 & 0.01 & 36.21
& 0.127 & 0.01 & 32.88 \\
\midrule

\multirow{2}{*}{Diffusion}
& UNet & Palette~\cite{Palette}
& 0.216 & 0.03 & 41.19
& 0.755 & 0.32 & 270.9
& 0.747 & 0.29 & 255.7
& 0.929 & 0.52 & 412.3 \\

& UNet & SelfRDB~\cite{selfRDB}
& 0.189 & 0.07 & 78.80
& 0.254 & 0.03 & 53.98
& 0.203 & 0.03 & 50.24
& 0.186 & 0.02 & 37.30 \\
\midrule

\multirow{4}{*}{Flow Matching}
& UNet & ConcatFM~\cite{lipman2023flow}
& 0.080 & 0.01 & 18.47
& 0.137 & \best{0.01} & \second{19.18}
& \second{0.128} & 0.02 & \second{18.44}
& 0.109 & 0.02 & 20.67 \\

& UNet & DirectFM~\cite{lipman2023flow}
& 0.074 & 0.01 & 15.08
& 0.214 & 0.03 & 26.74
& 0.201 & 0.03 & 28.83
& 0.169 & 0.02 & 26.41 \\

& UNet & PMRF~\cite{PMRF}
& 0.090 & 0.02 & 24.74
& 0.212 & 0.02 & 28.37
& 0.149 & 0.02 & 22.64
& 0.214 & 0.02 & 22.18 \\

\rowcolor{ourrow}
& UNet & \textbf{FlowMI (Ours)}
& \best{0.065} & 0.01 & 11.55
& \best{0.120} & \best{0.01} & \best{16.74}
& \best{0.101} & \best{0.01} & \best{14.63}
& \best{0.090} & 0.01 & \second{12.85} \\
\bottomrule
\end{tabular}
}
\vspace{-10pt}
\end{table*}

\subsection{FlowMI Implementation Details}

\noindent\textbf{Model Architecture.}
FlowMI consists of two stages: a 3D variational autoencoder (VAE) that maps volumes into a unified latent space, and a flow matching module that transports latents conditioned on the observation mask. Both stages use a 3D U-Net backbone with depth 4, residual blocks at each level, and group normalization. The VAE produces a latent representation with 256 channels; the flow module uses an identically structured U-Net to predict the continuous velocity field over latents. 

\noindent\textbf{Training Configuration.}
We train each stage with AdamW at a learning rate of $1\times10^{-4}$, using cosine annealing with linear warm-up over the first 5\% of steps. Due to the memory cost of 3D volumes, we use a per-GPU batch size of 2 with gradient accumulation, yielding an effective batch size of 8. All models are trained for 200 epochs unless otherwise specified. Following standard continuous-time flow matching, we sample $t \sim \mathcal{U}(0,1)$ uniformly at each training step. All experiments are run on NVIDIA A6000 GPUs. Notably, FlowMI uses a single shared model (one VAE and one flow module) across all three DCE tasks, whereas competing methods require a separate model for each setting.

\noindent\textbf{Metrics.} We evaluate synthesis quality at three levels. At the image level, we report 5 standard fidelity and perceptual metrics: Peak Signal-to-Noise Ratio (PSNR), Structural Similarity Index Measure (SSIM), Kernel Inception Distance (KID), Fréchet Inception Distance (FID), and Learned Perceptual Image Patch Similarity (LPIPS). At the lesion level, the tumor segmentation masks released with \ourbench{} enable measuring whether synthesized lesions remain faithful to the ground truth, reported via downstream segmentation performance. Finally, we conduct a reader study in which users subjectively assess image quality and diagnostic plausibility.

\subsection{Experiment Results}
\label{subsection:main results}

\begin{figure*}[t]
\begin{center}
\centerline{\includegraphics[width=1\linewidth]{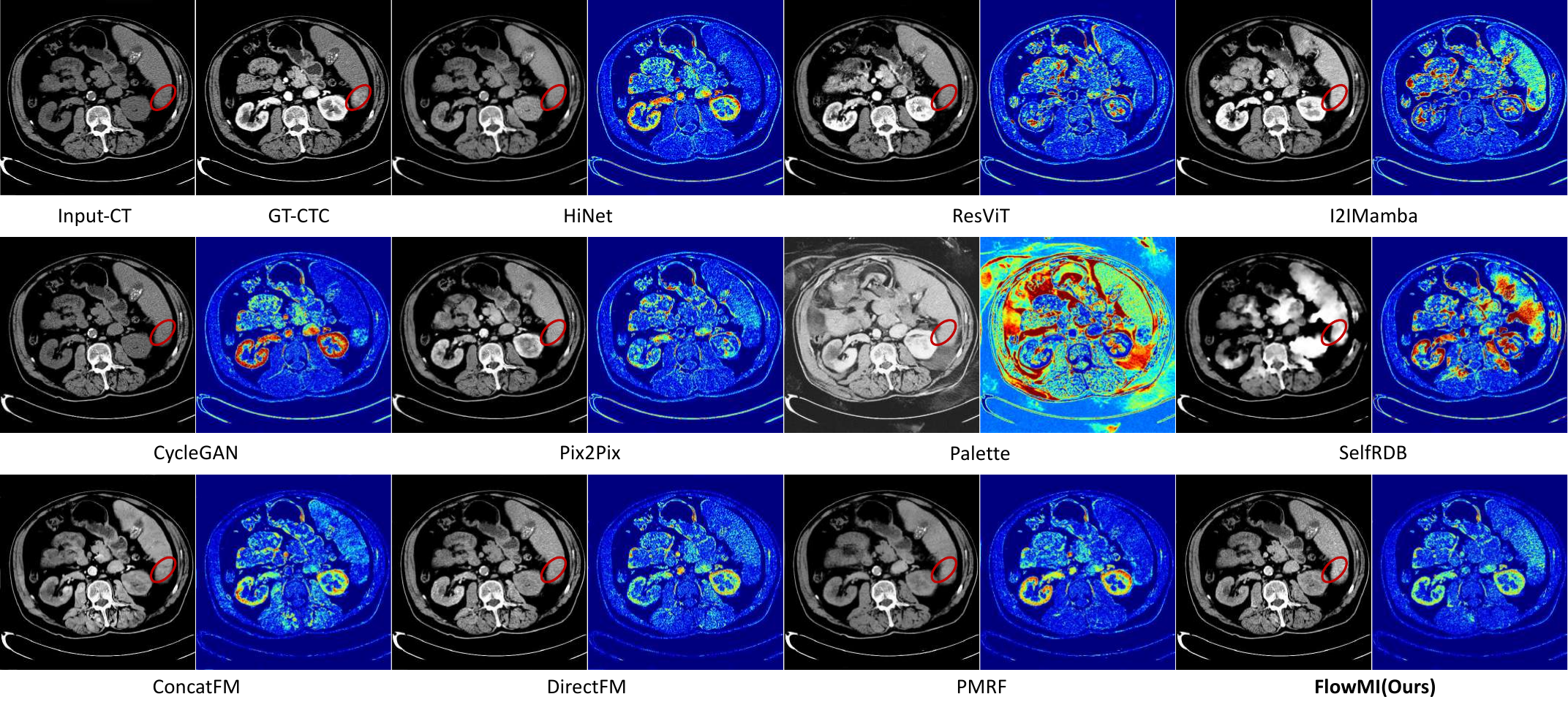}}
\vspace{-10pt}
\caption{
CT$\rightarrow$CTC (liver). Red circles mark tumor regions. 
} 
\label{fig:brain}
\vspace{-20pt}
\end{center}
\end{figure*}
\definecolor{gapred}{RGB}{180,40,40}

\begin{table}[h]
\centering
\scriptsize
\setlength{\tabcolsep}{6pt}
\caption{Downstream tumor segmentation performance. Dice scores are reported using nnU-Net trained on real CT (ground truth upper bound), with \textcolor{gapred}{red} indicating a deficit relative to ground truth. } 
\vspace{3pt}
\label{tb:seg_results}
\begin{tabular}{l c c c c c}
\toprule
\textbf{Metric} & \textbf{CycleGAN} & \textbf{I2IMamba} & \textbf{ResViT} & \textbf{Ours} & \textbf{Ground Truth} \\
\midrule
Dice $\uparrow$ & 0.61~\textcolor{gapred}{\scriptsize($-$0.02)} & 0.62~\textcolor{gapred}{\scriptsize($-$0.01)} & 0.60~\textcolor{gapred}{\scriptsize($-$0.03)} & \textbf{0.63}~\textcolor{gapred}{\scriptsize($\pm$0.00)} & \textbf{0.63} \\
\bottomrule
\end{tabular}
\vspace{-10pt}
\end{table}

\noindent\textbf{Quantitative comparison.}
Tables~\ref{tb:experiments_mechanism-based} and~\ref{tb:experiments_lpips_fid_kid} show that flow matching methods consistently outperform the direct, GAN, and diffusion baselines, with \ourmodel{} achieving the best PSNR, SSIM, and LPIPS, and competitive FID and KID across all tasks, with the largest margin on CT$\rightarrow$CTC. On the three DCE tasks, the gain over the second-best baseline (ConcatFM) is more modest in absolute terms but more meaningful in practice: ConcatFM and the other baselines train a separate model for each DCE setting, whereas \ourmodel{} handles all three settings with a single shared model.

\noindent\textbf{Qualitative comparison.}
Figure~\ref{fig:brain} reinforces the quantitative findings. Most baselines produce visually plausible CT images but leave tumor regions under-enhanced, making lesions hard to distinguish. \ourmodel{} instead reproduces the bright enhancement seen in the ground-truth CTC, and the residual maps (blue) confirm correspondingly lower reconstruction error in tumor regions.

\noindent\textbf{Downstream lesion analysis.}
Beyond image-level metrics, we evaluate whether the synthesized contrast scans preserve the lesion information needed for downstream analysis. We train a tumor segmentation model on real ground-truth contrast scans, then apply it to the synthesized outputs of each method and measure segmentation performance against the released tumor masks. The score on real ground-truth scans serves as an upper bound; the closer a method's score approaches this bound, the more faithfully its synthesized images preserve clinically actionable lesion structure. Table~\ref{tb:seg_results} reports Dice on the downstream segmentation task. FlowMI matches the upper bound in this setting and outperforms the other baselines.

\noindent\textbf{Leave-organs-out generalization.}
To test whether FlowMI learns a transferable contrast-enhancement operation, we hold out organs from training and evaluate on the held-out organ cases, comparing against the two strong baselines. We use four organs spanning tissue types: Colon, Ovary, Uterus, and Pancreas. As shown in Table~\ref{tb:leave-organ}, FlowMI achieves the highest PSNR and SSIM on every held-out organ, with the largest margin on Colon (+2.16 PSNR, +6.5 SSIM). The roughly 5 dB PSNR drop and higher standard deviation relative to the in-distribution setting (Table~\ref{tb:experiments_mechanism-based}) is expected for unseen anatomy, but FlowMI's consistent advantage across all four organs indicates that the latent transport generalizes rather than memorizes.  

\begin{table}[h]
\centering
\scriptsize
\setlength{\tabcolsep}{4pt}
\renewcommand{\arraystretch}{1.25}
\caption{Leave-organs-out generalization on the CT$\rightarrow$CTC task. Each column holds out one organ from training and evaluates on the held-out cases. PSNR (dB) and SSIM (\%) reported as Mean$\pm$Std.}
\label{tb:leave-organ}
\resizebox{\columnwidth}{!}{%
\begin{tabular}{l|cc|cc|cc|cc|cc}
\toprule
\multirow{2}{*}{\textbf{Method}}
& \multicolumn{2}{c|}{\textbf{Colon} (22)}
& \multicolumn{2}{c|}{\textbf{Ovary} (12)}
& \multicolumn{2}{c|}{\textbf{Uterus} (54)}
& \multicolumn{2}{c|}{\textbf{Pancreas} (54)}
& \multicolumn{2}{c}{\textbf{Average}} \\
\cmidrule(lr){2-3} \cmidrule(lr){4-5} \cmidrule(lr){6-7} \cmidrule(lr){8-9} \cmidrule(lr){10-11}
& PSNR & SSIM & PSNR & SSIM & PSNR & SSIM & PSNR & SSIM & PSNR & SSIM \\
\midrule
ConcatFM
& 17.12\,\tiny{$\pm$8.18} & 65.3\,\tiny{$\pm$15.4}
& 16.08\,\tiny{$\pm$8.27} & 59.1\,\tiny{$\pm$17.9}
& 18.94\,\tiny{$\pm$7.83} & 61.7\,\tiny{$\pm$13.6}
& 17.85\,\tiny{$\pm$7.49} & 62.2\,\tiny{$\pm$15.1}
& 17.50 & 62.1 \\
DirectFM
& 16.71\,\tiny{$\pm$8.42} & 58.9\,\tiny{$\pm$16.2}
& 17.63\,\tiny{$\pm$8.91} & 60.4\,\tiny{$\pm$16.8}
& 17.62\,\tiny{$\pm$7.07} & 59.8\,\tiny{$\pm$14.3}
& 17.46\,\tiny{$\pm$7.13} & 63.5\,\tiny{$\pm$14.7}
& 17.36 & 60.7 \\
\midrule
\rowcolor{ourrow}
\textbf{FlowMI (Ours)}
& \textbf{19.28\,\tiny{$\pm$7.71}} & \textbf{71.8\,\tiny{$\pm$10.8}}
& \textbf{18.94\,\tiny{$\pm$7.63}} & \textbf{63.7\,\tiny{$\pm$15.4}}
& \textbf{20.41\,\tiny{$\pm$6.52}} & \textbf{67.9\,\tiny{$\pm$12.9}}
& \textbf{19.87\,\tiny{$\pm$6.89}} & \textbf{66.2\,\tiny{$\pm$13.5}}
& \textbf{19.63} & \textbf{67.4} \\
\bottomrule
\end{tabular}%
}
\end{table}


\noindent\textbf{Reader Study.}
Image-quality metrics correlate imperfectly with diagnostic utility, so we complement them with a reader study evaluating clinical plausibility. We recruited 8 experienced readers with backgrounds in radiology and medical imaging research to evaluate the synthesized scans across three complementary tasks. In the \textit{authenticity classification} task, readers are shown a single scan (real or synthesized, presented in randomized order) and asked to judge whether it is real; an ideal synthesis method drives accuracy toward chance (50\%). In the \textit{quality scoring} task, readers rate each synthesized scan on a 5-point Likert scale for overall clinical realism. In the \textit{paired preference} task, readers are shown two synthesized scans from different methods side by side and select which is more realistic; we report the win rate of each method against all others. Each reader evaluates 34 cases sampled uniformly across organs, with case order randomized and method labels blinded. Table~\ref{tb:reader_study} summarizes the results. FlowMI achieves the lowest authenticity-classification accuracy (56.3\%, closest to the 50\% ideal), the highest mean Likert score (3.99), and the highest paired-preference win rate (72.5\%), indicating that radiologists find its outputs the hardest to distinguish from real scans and the most clinically plausible.

\begin{table}[h]
\centering
\small
\caption{Reader study results across three tasks. \textbf{Authenticity}: classification accuracy on real-vs-synthesized judgments (closer to 50\% is better; \textcolor{gapred}{red} indicates deviation from the 50\% ideal). \textbf{Likert}: mean rating on a 1--5 scale for clinical realism (higher is better). \textbf{Preference}: pairwise win rate against all other methods (higher is better). Values are mean~$\pm$~std across readers. Best results in \textbf{bold}.}
\label{tb:reader_study}
\resizebox{0.5\columnwidth}{!}{%
\begin{tabular}{l c c c}
\toprule
& \textbf{Authenticity} & \textbf{Likert} & \textbf{Preference} \\
\cmidrule(lr){2-2} \cmidrule(lr){3-3} \cmidrule(lr){4-4}
\textbf{Method} & \textbf{Acc.\ (\%)} $\rightarrow 50$ & \textbf{Score} $\uparrow$ & \textbf{Win Rate (\%)} $\uparrow$ \\
\midrule
CycleGAN   & 68.4$_{\pm 14.7}$~\textcolor{gapred}{\scriptsize($+$18.4)} & 3.58$_{\pm 0.63}$ & 69.1$_{\pm 33.2}$ \\
I2IMamba   & 61.2$_{\pm 15.3}$~\textcolor{gapred}{\scriptsize($+$11.2)} & 3.05$_{\pm 0.69}$ & 58.7$_{\pm 34.7}$ \\
ResViT     & 64.6$_{\pm 20.1}$~\textcolor{gapred}{\scriptsize($+$14.6)} & 3.34$_{\pm 0.74}$ & 65.3$_{\pm 23.8}$ \\
\midrule
\rowcolor{ourrow} \textbf{FlowMI (Ours)} & \textbf{56.3}$_{\pm 15.6}$~\textcolor{gapred}{\scriptsize($+$6.3)} & \textbf{3.99}$_{\pm 0.79}$ & \textbf{72.5}$_{\pm 20.5}$ \\
\midrule
Real (GT)  & --- & 4.60$_{\pm 0.34}$ & --- \\
\bottomrule
\end{tabular}%
}
\end{table}

\smallskip\smallskip\noindent\textbf{Ablation Study.}
We ablate three components of FlowMI: (i) the trajectory reconstruction loss $\mathcal{L}_{\text{Trajectory}}$ (Eq.~\ref{eq:rec_loss}), which anchors intermediate latents to the image-space interpolant; (ii) mask conditioning, which signals the observation pattern to the velocity field; and (iii) Product-of-Experts (PoE) fusion, replaced with mean-pooling over modality-specific latents. As shown in Table~\ref{tb:ablation}, removing $\mathcal{L}_{\text{Trajectory}}$ causes the largest drop (\(-1.49\) avg.~PSNR), confirming that trajectory anchoring is critical across both CT and DCE settings. Mask conditioning and PoE have negligible effect on single-modality CT$\rightarrow$CTC (within $\pm0.05$ dB), as expected since both reduce to near-identity operations with one input. Their value emerges on the multi-modality DCE tasks, where mask conditioning lets the velocity field distinguish $1{\rightarrow}n$ from $n{\rightarrow}1$ configurations and PoE's precision-weighted fusion outperforms uniform averaging, costing \(-0.88\) and \(-0.33\) avg.~PSNR respectively when removed.

\begin{table*}[h]
\centering
\scriptsize
\setlength{\tabcolsep}{4pt}
\renewcommand{\arraystretch}{1.25}
\caption{Ablation of FlowMI components. PSNR (dB) and SSIM (\%) reported as Mean$\pm$Std on the four tasks. Average columns report the unweighted mean across the four tasks.}
\label{tb:ablation}
\resizebox{\textwidth}{!}{%
\begin{tabular}{l|cc|cc|cc|cc|cc}
\toprule
\multirow{2}{*}{\textbf{Variant}}
& \multicolumn{2}{c|}{\textbf{CT$\rightarrow$CTC}}
& \multicolumn{2}{c|}{\textbf{DCE$_1\!\rightarrow$DCE$_2$}}
& \multicolumn{2}{c|}{\textbf{DCE$_1\!\rightarrow$DCE$_{2,3}$}}
& \multicolumn{2}{c|}{\textbf{DCE$_{1,3}\!\rightarrow$DCE$_2$}}
& \multicolumn{2}{c}{\textbf{Average}} \\
\cmidrule(lr){2-3} \cmidrule(lr){4-5} \cmidrule(lr){6-7} \cmidrule(lr){8-9} \cmidrule(lr){10-11}
& PSNR & SSIM & PSNR & SSIM & PSNR & SSIM & PSNR & SSIM & PSNR & SSIM \\
\midrule
w/o $\mathcal{L}_{\text{Trajectory}}$
& 22.81\,\tiny{$\pm$4.32} & 75.2\,\tiny{$\pm$9.74}
& 24.97\,\tiny{$\pm$3.41} & 70.4\,\tiny{$\pm$10.2}
& 25.18\,\tiny{$\pm$3.38} & 70.1\,\tiny{$\pm$8.96}
& 27.89\,\tiny{$\pm$3.52} & 73.4\,\tiny{$\pm$8.41}
& 25.21 & 72.3 \\
w/o mask conditioning
& 24.49\,\tiny{$\pm$4.18} & 78.4\,\tiny{$\pm$8.83}
& 25.73\,\tiny{$\pm$3.27} & 71.8\,\tiny{$\pm$9.65}
& 25.42\,\tiny{$\pm$3.24} & 70.6\,\tiny{$\pm$8.72}
& 27.64\,\tiny{$\pm$3.41} & 72.1\,\tiny{$\pm$8.27}
& 25.82 & 73.2 \\
PoE $\rightarrow$ mean-pooling
& 24.46\,\tiny{$\pm$4.21} & 78.1\,\tiny{$\pm$9.01}
& 26.08\,\tiny{$\pm$3.19} & 73.1\,\tiny{$\pm$9.48}
& 26.21\,\tiny{$\pm$3.16} & 72.5\,\tiny{$\pm$8.34}
& 28.74\,\tiny{$\pm$3.31} & 75.1\,\tiny{$\pm$7.92}
& 26.37 & 74.7 \\
\midrule
\rowcolor{ourrow}
\textbf{FlowMI (full)}
& \textbf{24.47\,\tiny{$\pm$4.15}} & \textbf{78.5\,\tiny{$\pm$8.62}}
& \textbf{26.52\,\tiny{$\pm$3.13}} & \textbf{74.2\,\tiny{$\pm$9.33}}
& \textbf{26.63\,\tiny{$\pm$3.11}} & \textbf{73.7\,\tiny{$\pm$8.06}}
& \textbf{29.17\,\tiny{$\pm$3.24}} & \textbf{76.2\,\tiny{$\pm$7.65}}
& \textbf{26.70} & \textbf{75.7} \\
\bottomrule
\end{tabular}%
}
\end{table*}

\section{Conclusion}

In this work, we introduced \ourdataset{}, a comprehensive pan-cancer benchmark for multimodal image translation and missing-modality synthesis in clinically realistic settings. 
It provides high-quality, per-patient paired CT and MR scans across 11 organs with both contrast-enhanced and non-contrast modalities, curated through a standardized and reproducible pipeline. 
We defined three benchmark tasks and reported reference results using representative generative models. 
While limitations remain, such as biases in public data sources and variability in clinical imaging, \ourdataset{} offers a foundation for developing robust multimodal translation methods and advancing clinically reliable decision support. 
Future work could extend the dataset with additional modalities, larger cohorts, and more advanced synthesis techniques to further enhance generalization and applicability.

{\small
\bibliographystyle{splncs04}
\bibliography{main}

@String(ICCV= {Int. Conf. Comput. Vis.})

@String(ICCV  = {ICCV})

@article{chen2025learning,
  title={Learning Patient-Specific Disease Dynamics with Latent Flow Matching for Longitudinal Imaging Generation},
  author={Chen, Hao and Yin, Rui and Chen, Yifan and Chen, Qi and Li, Chao},
  journal={arXiv preprint arXiv:2512.09185},
  year={2025}
}

@misc{TCIA,
  author = {Kirk, S. and Lee, Y. and Sadow, C. A. and Levine, S.},
  title = {The Cancer Genome Atlas Rectum Adenocarcinoma Collection (TCGA-READ)},
  year = {2016},
  note = {Version 3 [Dataset]. The Cancer Imaging Archive. \url{https://doi.org/10.7937/K9/TCIA.2016.F7PPNPNU}},
  doi = {10.7937/K9/TCIA.2016.F7PPNPNU}
}

@article{TCGA-BRCA,
  title={The cancer genome atlas breast invasive carcinoma collection (TCGA-BRCA)},
  author={Lingle, Wilma and Erickson, Bradley J and Zuley, Margarita L and Jarosz, Rose and Bonaccio, Ermelinda and Filippini, Joe and Net, Jose M and Levi, Len and Morris, Elizabeth A and Figler, Gloria G and others},
  journal={(No Title)},
  year={2016},
  publisher={The Cancer Imaging Archive}
}

@article{Adrenal-ACC-Ki67-Seg,
  title={Voxel-level segmentation of pathologically-proven Adrenocortical carcinoma with Ki-67 expression (Adrenal-ACC-Ki67-Seg)[Data set]},
  author={Moawad, Ahmed W and Ahmed, Ayahallah A and ElMohr, Mohab and Eltaher, Mohamed and Habra, Mouhammed Amir and Fisher, Sarah and Perrier, Nancy and Zhang, Miao and Fuentes, David and Elsayes, Khaled},
  journal={The Cancer Imaging Archive},
  volume={8},
  year={2023}
}

@article{TCGA-OV,
  title={The cancer genome atlas ovarian cancer collection (tcga-ov)(version 4)[Dataset]},
  author={Holback, Chandra and Jarosz, Rose and Prior, Fred and Mutch, David G and Bhosale, Priya and Garcia, Kimberly and Lee, Yueh and Kirk, Shanah and Sadow, Cheryl A and Levine, Seth and others},
  journal={The Cancer Imaging Archive},
  volume={10},
  pages={K9},
  year={2016}
}

@article{TCGA-UCEC,
  title={The cancer genome atlas uterine corpus endometrial carcinoma collection (tcga-ucec)},
  author={Erickson, BJ and Mutch, D and Lippmann, L and Jarosz, R},
  journal={The Cancer Imaging Archive},
  year={2016}
}

@misc{CPTAC-UCEC,
  author = {{National Cancer Institute Clinical Proteomic Tumor Analysis Consortium (CPTAC)}},
  title  = {The Clinical Proteomic Tumor Analysis Consortium Uterine Corpus Endometrial Carcinoma Collection (CPTAC-UCEC) (Version 10) [Dataset]},
  year   = {2019},
  note   = {The Cancer Imaging Archive.}
}

@article{TCGA-STAD,
  title={The cancer genome atlas stomach adenocarcinoma collection (TCGA-STAD)},
  author={Lucchesi, FR and Aredes, ND},
  journal={The Cancer Imaging Archive},
  year={2016}
}

@misc{CPTAC-PDA,
  author = {{National Cancer Institute Clinical Proteomic Tumor Analysis Consortium (CPTAC)}},
  title  = {The Clinical Proteomic Tumor Analysis Consortium Pancreatic Ductal Adenocarcinoma Collection (CPTAC-PDA) (Version 15) [Dataset]},
  year   = {2018},
  note   = {The Cancer Imaging Archive. \url{https://doi.org/10.7937/K9/TCIA.2018.SC20FO18}}
}

@article{HCC-TACE-Seg,
  title={Multimodality annotated HCC cases with and without advanced imaging segmentation},
  author={Moawad, AW and Fuentes, D and Morshid, A and Khalaf, AM and Elmohr, MM and Abusaif, A and Hazle, JD and Kaseb, AO and Hassan, M and Mahvash, A and others},
  journal={The Cancer Imaging Archive (TCIA)},
  year={2021}
}

@article{TCGA-LIHC,
  title={The Cancer genome atlas liver hepatocellular carcinoma collection (TCGA-LIHC)(Version 5)[Dataset]},
  author={Erickson, Bradley J and Kirk, Shanah and Lee, Y and Bathe, Oliver and Kearns, Melissa and Gerdes, C and Rieger-Christ, Kimberly and Lemmerman, John},
  journal={The Cancer Imaging Archive},
  year={2016}
}

@misc{CMB-CRC,
  author = {Cancer Moonshot Biobank},
  title  = {Cancer Moonshot Biobank – Colorectal Cancer Collection (CMB-CRC) (Version 8) [Dataset]},
  note   = {The Cancer Imaging Archive. \url{https://doi.org/10.7937/djg7-gz87}},
  year   = 2022
}

@article{TCGA-COAD,
  title={The Cancer Genome Atlas Colon Adenocarcinoma Collection (TCGA-COAD)(Version 3)[Dataset]},
  author={Kirk, S and Lee, Y and Sadow, CA and Levine, S and Roche, C and Bonaccio, E and Filiippini, J},
  journal={The Cancer Imaging Archive. https://doi. org/10.7937 K},
  volume={9},
  year={2016}
}

@incollection{TCGA-BLCA,
  title={The cancer genome atlas urothelial bladder carcinoma collection (TCGA-BLCA)},
  author={Kirk, Shanah and Lee, Yueh and Lucchesi, Fabiano R and Aredes, Natalia D and Gruszauskas, Nicholas and Catto, James and Garcia, Kimberly and Jarosz, Rose and Duddalwar, Vinay and Varghese, Bino and others},
  booktitle={The Cancer Imaging Archive},
  year={2016}
}

@article{TCGA-KIRC,
  title={The cancer genome atlas kidney renal clear cell carcinoma collection (TCGA-KIRC)(Version 3)[Dataset]},
  author={Akin, Oguz and Elnajjar, Pierre and Heller, Matthew and Jarosz, Rose and Erickson, Bradley J and Kirk, Shanah and Lee, Yueh and Linehan, Marston W and Gautam, Rabindra and Vikram, Raghu and others},
  journal={Cancer Imaging Arch},
  year={2016}
}

@misc{C4KC-KiTS,
  author = {Nicholas Heller and Nithesh Sathianathen and Arveen Kalapara and Ethan Walczak and Kenneth Moore and Holly Kaluzniak and Jacob Rosenberg and Paul Blake and Zachary Rengel and Michael Oestreich and Joel Dean and Matthew Tradewell and Adeel Shah and Rishi Tejpaul and Zachary Edgerton and Matthew Peterson and Sohaib Raza and Samip Regmi and Nikolaos Papanikolopoulos and Christopher Weight},
  title  = {Data from C4KC-KiTS [Dataset]},
  note   = {The Cancer Imaging Archive. \url{https://doi.org/10.7937/TCIA.2019.IX49E8NX}},
  year   = 2019
}

@misc{CPTAC-CCRCC,
  author = {{National Cancer Institute Clinical Proteomic Tumor Analysis Consortium (CPTAC)}},
  title  = {The Clinical Proteomic Tumor Analysis Consortium Clear Cell Renal Cell Carcinoma Collection (CPTAC-CCRCC)},
  note   = {The Cancer Imaging Archive. \url{https://doi.org/10.7937/k9/tcia.2018.oblamn27}},
  year   = 2018
}

@article{TCGA-KIRP,
  title={The cancer genome atlas cervical kidney renal papillary cell carcinoma collection (TCGA-KIRP), version 4},
  author={Linehan, Marston and Gautam, R and Kirk, S and Lee, Y and Roche, C and Bonaccio, E and Filippini, J and Rieger-Christ, K and Lemmerman, J and Jarosz, R},
  journal={The Cancer Imaging Archive},
  year={2016}
}

@article{TCGA-KICH,
  title={The Cancer Genome Atlas Kidney Chromophobe Collection (TCGA-KICH)(Version 3)},
  author={Linehan, M and Gautam, R and Sadow, C and Levine, SJ},
  journal={The Cancer Imaging Archive},
  year={2016}
}

@article{CMB-LCA,
  title={Cancer moonshot biobank-lung cancer collection (cmb-lca)(version 3)[dataset]},
  author={Biobank, C},
  journal={The Cancer Imaging Archive},
  year={2022}
}

@misc{Lung-PET-CT-Dx,
  author = {Peng Li and Shaoke Wang and Tianyu Li and Jie Lu and Yufan HuangFu and Dong Wei Wang},
  title  = {A Large-Scale CT and PET/CT Dataset for Lung Cancer Diagnosis (Lung-PET-CT-Dx) [Dataset]},
  note   = {The Cancer Imaging Archive. \url{https://doi.org/10.7937/TCIA.2020.NNC2-0461}},
  year   = 2020
}

@misc{Anti-PD-1_Lung,
  author = {Pranathi Madhavi and Shweta Patel and Anne S. Tsao},
  title  = {Data from Anti-PD-1 Immunotherapy Lung [Dataset]},
  note   = {The Cancer Imaging Archive. \url{https://doi.org/10.7937/tcia.2019.zjjwb9ip}},
  year   = 2019
}

@article{TCGA-LUSC,
  title={The cancer genome atlas lung squamous cell carcinoma collection (tcga-lusc)(version 4)[Dataset]},
  author={Kirk, S and Lee, Y and Kumar, P and Filippini, J and Albertina, B and Watson, M and Rieger-Christ, K and Lemmerman, J},
  journal={The Cancer Imaging Archive},
  year={2016}
}

@article{CPTAC-LSCC,
  title={The clinical proteomic tumor analysis consortium lung squamous cell carcinoma collection (CPTAC-LSCC)},
  author={{National Cancer Institute Clinical Proteomic Tumor Analysis Consortium and others}},
  journal={The Cancer Imaging Archive},
  year={2018},
  publisher={The Cancer Imaging Archive}
}

@article{CPTAC-LUAD,
  title={The clinical proteomic tumor analysis consortium lung adenocarcinoma collection (CPTAC-LUAD)},
  author={{National Cancer Institute Clinical Proteomic Tumor Analysis Consortium and others}},
  journal={The Cancer Imaging Archive.},
  year={2018}
}

@article{ucsf1,
  title={Invasive breast cancer: predicting disease recurrence by using high-spatial-resolution signal enhancement ratio imaging},
  author={Li, Ka-Loh and Partridge, Savannah C and Joe, Bonnie N and Gibbs, Jessica E and Lu, Ying and Esserman, Laura J and Hylton, Nola M},
  journal={Radiology},
  volume={248},
  number={1},
  pages={79--87},
  year={2008},
  publisher={Radiological Society of North America}
}

@article{ucsf2,
  title={Optimized breast MRI functional tumor volume as a biomarker of recurrence-free survival following neoadjuvant chemotherapy},
  author={Jafri, Nazia F and Newitt, David C and Kornak, John and Esserman, Laura J and Joe, Bonnie N and Hylton, Nola M},
  journal={Journal of Magnetic Resonance Imaging},
  volume={40},
  number={2},
  pages={476--482},
  year={2014},
  publisher={Wiley Online Library}
}

@article{I-SPY1,
  title={Multi-center breast DCE-MRI data and segmentations from patients in the I-SPY 1/ACRIN 6657 trials},
  author={Newitt, David and Hylton, Nola and others},
  journal={Cancer Imaging Arch},
  volume={10},
  number={7},
  pages={2016},
  year={2016}
}

@article{Breast3,
  title={Radiological tumour classification across imaging modality and histology},
  author={Wu, Jia and Li, Chao and Gensheimer, Michael and Padda, Sukhmani and Kato, Fumi and Shirato, Hiroki and Wei, Yiran and Sch{\"o}nlieb, Carola-Bibiane and Price, Stephen John and Jaffray, David and others},
  journal={Nature machine intelligence},
  volume={3},
  number={9},
  pages={787--798},
  year={2021},
  publisher={Nature Publishing Group UK London}
}

@inproceedings{CHAOS,
  title={Multi-modal learning from unpaired images: Application to multi-organ segmentation in CT and MRI},
  author={Valindria, Vanya V and Pawlowski, Nick and Rajchl, Martin and Lavdas, Ioannis and Aboagye, Eric O and Rockall, Andrea G and Rueckert, Daniel and Glocker, Ben},
  booktitle={2018 IEEE winter conference on applications of computer vision (WACV)},
  pages={547--556},
  year={2018},
  organization={IEEE},
  note = {\url{}}
}

@article{SynthRAD2023,
  title={SynthRAD2023 Grand Challenge dataset: Generating synthetic CT for radiotherapy},
  author={Thummerer, Adrian and Van der Bijl, Erik and Galapon Jr, Arthur and Verhoeff, Joost JC and Langendijk, Johannes A and Both, Stefan and van den Berg, Cornelis (Nico) AT and Maspero, Matteo},
  journal={Medical physics},
  volume={50},
  number={7},
  pages={4664--4674},
  year={2023},
  publisher={Wiley Online Library},
  note = {\url{https://ieeexplore.ieee.org/document/8354170}}
}

@article{BraTS2024,
  title        = {The 2024 brain tumor segmentation (brats) challenge: Glioma segmentation on post-treatment mri},
  author       = {de Verdier, Maria Correia and Saluja, Rachit and Gagnon, Louis and LaBella, Dominic and Baid, Ujjwall and Tahon, Nourel Hoda and Foltyn-Dumitru, Martha and Zhang, Jikai and Alafif, Maram and Baig, Saif and others},
  journal      = {arXiv preprint arXiv:2405.18368},
  year         = {2024},
  note   = {\url{https://arxiv.org/abs/2405.18368}}
}

@article{BraSyn,
  title={The brain tumor segmentation (BraTS) challenge 2023: Brain MR image synthesis for tumor segmentation (BraSyn)},
  author={Li, Hongwei Bran and Conte, Gian Marco and Hu, Qingqiao and Anwar, Syed Muhammad and Kofler, Florian and Ezhov, Ivan and van Leemput, Koen and Piraud, Marie and Diaz, Maria and Cole, Byrone and others},
  journal={ arXiv preprint arXiv:2305.09011},
  year={2024},
  note = {\url{https://pmc.ncbi.nlm.nih.gov/articles/PMC10441440/}}
}

@misc{IXI,
  author       = {{Biomedical Image Analysis Group, Imperial College London}},
  title        = {{IXI} {D}ataset -- {I}nformation e{X}traction from {I}mages},
  howpublished = {\url{https://brain-development.org/ixi-dataset/}},
  note         = {EPSRC GR/S21533/02. RRID:SCR\_005839. Accessed: 2026-05-07},
}

@article{OASIS-3,
  title={OASIS-3: Longitudinal Neuroimaging, Clinical, and Cognitive Dataset for Normal Aging and Alzheimer Disease},
  author={LaMontagne, Pamela J and Benzinger, Tammie L. S. and Morris, John C and Keefe, Sarah and Hornbeck, Russ and Xiong, Chengjie and Grant, Elizabeth and Hassenstab, Jason and Moulder, Krista and Vlassenko, Andrei G. and others},
  journal={medRxiv},
  year={2019},
  note={\url{https://doi.org/10.1101/2019.12.13.19014902}}
}

@article{ADNI-3,
  title={The Alzheimer's Disease Neuroimaging Initiative (ADNI): MRI Methods},
  author={Jack, Clifford R. Jr. and Bernstein, Matt A. and Fox, Nick C. and Thompson, Paul and Alexander, Gene and Harvey, Danielle and Borowski, Bret and Britson, Paula J. and Whitwell, Jennifer L. and Ward, Chad and others},
  journal={Journal of Magnetic Resonance Imaging},
  volume={27},
  number={4},
  pages={685--691},
  year={2008},
  note={\url{https://doi.org/10.1002/jmri.21049}}
}

@article{AANLIB,
  title={Harvard Whole Brain Atlas: www. med. harvard. edu/AANLIB/home. html},
  author={Summers, D},
  journal={Journal of Neurology, Neurosurgery \& Psychiatry},
  volume={74},
  number={3},
  pages={288--288},
  year={2003},
  publisher={BMJ Publishing Group Ltd},
  note = {\url{https://jnnp.bmj.com/content/74/3/288}}
}

@article{CrossMoDA,
  title={CrossMoDA 2021 challenge: Benchmark of cross-modality domain adaptation techniques for vestibular schwannoma and cochlea segmentation},
  author={Dorent, Reuben and Kujawa, Aaron and Ivory, Marina and Bakas, Spyridon and Rieke, Nicola and Joutard, Samuel and Glocker, Ben and Cardoso, Jorge and Modat, Marc and Batmanghelich, Kayhan and others},
  journal={Medical Image Analysis},
  volume={83},
  pages={102628},
  year={2023},
  publisher={Elsevier},
  note = {\url{https://www.sciencedirect.com/science/article/pii/S1361841522002560}}
}

@article{ACDC,
  title={Deep learning techniques for automatic MRI cardiac multi-structures segmentation and diagnosis: is the problem solved?},
  author={Bernard, Olivier and Lalande, Alain and Zotti, Clement and Cervenansky, Frederick and Yang, Xin and Heng, Pheng-Ann and Cetin, Irem and Lekadir, Karim and Camara, Oscar and Ballester, Miguel Angel Gonzalez and others},
  journal={IEEE transactions on medical imaging},
  volume={37},
  number={11},
  pages={2514--2525},
  year={2018},
  publisher={ieee},
  note = {\url{https://ieeexplore.ieee.org/document/8360453}}
}

@article{MMWHS,
  title={Multivariate mixture model for myocardial segmentation combining multi-source images},
  author={Zhuang, Xiahai},
  journal={IEEE transactions on pattern analysis and machine intelligence},
  volume={41},
  number={12},
  pages={2933--2946},
  year={2018},
  publisher={IEEE},
  note = {\url{https://ieeexplore.ieee.org/abstract/document/8458220}}
}

@article{FDG-PET/CT,
  title={A whole-body FDG-PET/CT dataset with manually annotated tumor lesions},
  author={Gatidis, Sergios and Hepp, Tobias and Fr{\"u}h, Marcel and La Foug{\`e}re, Christian and Nikolaou, Konstantin and Pfannenberg, Christina and Sch{\"o}lkopf, Bernhard and K{\"u}stner, Thomas and Cyran, Clemens and Rubin, Daniel},
  journal={Scientific Data},
  volume={9},
  number={1},
  pages={601},
  year={2022},
  publisher={Nature Publishing Group UK London},
  note = {\url{https://www.scopus.com/pages/publications/85139504476}}
}

@misc{UDPET,
  title        = {Ultra-Low Dose PET Imaging Challenge 2024 (UDPET)},
  howpublished = {Dataset and challenge information available online},
  year         = {2024},
  note         = {Accessed via \url{https://udpet-challenge.github.io/}},
}

@article{HiNet,
  title={Hi-net: hybrid-fusion network for multi-modal MR image synthesis},
  author={Zhou, Tao and Fu, Huazhu and Chen, Geng and Shen, Jianbing and Shao, Ling},
  journal={IEEE transactions on medical imaging},
  volume={39},
  number={9},
  pages={2772--2781},
  year={2020},
  publisher={IEEE}
}

@inproceedings{CycleGAN,
  title={Unpaired image-to-image translation using cycle-consistent adversarial networks},
  author={Zhu, Jun-Yan and Park, Taesung and Isola, Phillip and Efros, Alexei A},
  booktitle={Proceedings of the IEEE international conference on computer vision},
  pages={2223--2232},
  year={2017}
}

@inproceedings{Pix2Pix,
  title={Image-to-image translation with conditional adversarial networks},
  author={Isola, Phillip and Zhu, Jun-Yan and Zhou, Tinghui and Efros, Alexei A},
  booktitle={Proceedings of the IEEE conference on computer vision and pattern recognition},
  pages={1125--1134},
  year={2017}
}

@article{ResViT,
  title={ResViT: Residual vision transformers for multimodal medical image synthesis},
  author={Dalmaz, Onat and Yurt, Mahmut and {\c{C}}ukur, Tolga},
  journal={IEEE Transactions on Medical Imaging},
  volume={41},
  number={10},
  pages={2598--2614},
  year={2022},
  publisher={IEEE}
}

@article{I2I,
  title={I2I-Mamba: Multi-modal medical image synthesis via selective state space modeling},
  author={Atli, Omer F and Kabas, Bilal and Arslan, Fuat and Demirtas, Arda C and Yurt, Mahmut and Dalmaz, Onat and Cukur, Tolga},
  journal={arXiv preprint arXiv:2405.14022},
  year={2024}
}

@article{selfRDB,
  title={Self-consistent recursive diffusion bridge for medical image translation},
  author={Arslan, Fuat and Kabas, Bilal and Dalmaz, Onat and Ozbey, Muzaffer and {\c{C}}ukur, Tolga},
  journal={Medical Image Analysis},
  volume={106},
  pages={103747},
  year={2025},
  publisher={Elsevier}
}

@inproceedings{Palette,
  title={Palette: Image-to-image diffusion models},
  author={Saharia, Chitwan and Chan, William and Chang, Huiwen and Lee, Chris and Ho, Jonathan and Salimans, Tim and Fleet, David and Norouzi, Mohammad},
  booktitle={ACM SIGGRAPH 2022 conference proceedings},
  pages={1--10},
  year={2022}
}

@article{PMRF,
  title={Posterior-mean rectified flow: Towards minimum mse photo-realistic image restoration},
  author={Ohayon, Guy and Michaeli, Tomer and Elad, Michael},
  journal={arXiv preprint arXiv:2410.00418},
  year={2024}
}

@inproceedings{LatentDiff,
  title={High-resolution image synthesis with latent diffusion models},
  author={Rombach, Robin and Blattmann, Andreas and Lorenz, Dominik and Esser, Patrick and Ommer, Bj{\"o}rn},
  booktitle={Proceedings of the IEEE/CVF conference on computer vision and pattern recognition},
  pages={10684--10695},
  year={2022}
}

@inproceedings{DiT,
  title={Scalable diffusion models with transformers},
  author={Peebles, William and Xie, Saining},
  booktitle=ICCV,
  pages={4195--4205},
  year={2023}
}

@inproceedings{liu2022flow,
  title     = {Flow Straight and Fast: Learning to Generate and Transfer Data with Rectified Flow},
  author    = {Liu, Xingchao and Gong, Chengyue and others},
  booktitle = {The Eleventh International Conference on Learning Representations},
  year      = {2022}
}

@inproceedings{lipman2023flow,
  title     = {Flow Matching for Generative Modeling},
  author    = {Lipman, Yaron and Chen, Ricky TQ and Ben-Hamu, Heli and Nickel, Maximilian and Le, Matthew},
  booktitle = {The Eleventh International Conference on Learning Representations},
  year      = {2023}
}

@article{xu2024diffusion,
  title={Diffusion models trained with large data are transferable visual models},
  author={Xu, Guangkai and Ge, Yongtao and Liu, Mingyu and Fan, Chengxiang and Xie, Kangyang and Zhao, Zhiyue and Chen, Hao and Shen, Chunhua},
  journal={arXiv e-prints},
  pages={arXiv--2403},
  year={2024}
}

@article{Chartsias2018,
   author = {Agisilaos Chartsias and Thomas Joyce and Mario Valerio Giuffrida and Sotirios A. Tsaftaris},
   doi = {10.1109/TMI.2017.2764326},
   issn = {1558254X},
   issue = {3},
   journal = {IEEE Transactions on Medical Imaging},
   month = {3},
   pages = {803-814},
   pmid = {29053447},
   publisher = {Institute of Electrical and Electronics Engineers Inc.},
   title = {Multimodal MR Synthesis via Modality-Invariant Latent Representation},
   volume = {37},
   year = {2018},
}

@article{dao2023flow,
  title={Flow matching in latent space},
  author={Dao, Quan and Phung, Hao and Nguyen, Binh and Tran, Anh},
  journal={arXiv preprint arXiv:2307.08698},
  year={2023}
}

@article{chadebec2025lbm,
  title={LBM: Latent Bridge Matching for Fast Image-to-Image Translation},
  author={Chadebec, Cl{\'e}ment and Tasar, Onur and Sreetharan, Sanjeev and Aubin, Benjamin},
  journal={arXiv preprint arXiv:2503.07535},
  year={2025}
}

@article{ji2022amos_dataset,
  title={Amos: A large-scale abdominal multi-organ benchmark for versatile medical image segmentation},
  author={Ji, Yuanfeng and Bai, Haotian and Ge, Chongjian and Yang, Jie and Zhu, Ye and Zhang, Ruimao and Li, Zhen and Zhanng, Lingyan and Ma, Wanling and Wan, Xiang and others},
  journal={Advances in neural information processing systems},
  volume={35},
  pages={36722--36732},
  year={2022}
}

@Article{Klein2010Elastix,
  author  = {Stefan Klein and Margreet Staring and Koen Murphy and et al.},
  title   = {elastix: a toolbox for intensity‐based medical image registration},
  journal = {IEEE Transactions on Medical Imaging},
  year    = {2010},
  volume  = {29},
  number  = {1},
  pages   = {196--205},
  doi     = {10.1109/TMI.2009.2035616}
}

@inproceedings{wu2018mvae,
  title     = {Multimodal Generative Models for Scalable Weakly-Supervised Learning},
  author    = {Wu, Mike and Goodman, Noah},
  booktitle = {Advances in Neural Information Processing Systems},
  volume    = {31},
  year      = {2018},
  note      = {\url{https://papers.neurips.cc/paper/7801-multimodal-generative-models-for-scalable-weakly-supervised-learning}}
}
}

\clearpage
\vspace{2em}

\noindent {\Large{Appendix}}

\appendix



\section{Additional Dataset Details}
Here we provide extended statistics for \ourdataset{} as well as the original data sources of each collection.  

All scans were obtained by downloading raw DICOM or NIfTI files from publicly available repositories. 
The included collections are:  
Adrenal-ACC-Ki67-Seg~\cite{Adrenal-ACC-Ki67-Seg}
TCGA-OV \cite{TCGA-OV}, 
TCGA-UCEC \cite{TCGA-UCEC}, 
CPTAC-UCEC \cite{CPTAC-UCEC}, 
TCGA-STAD \cite{TCGA-STAD}, 
CPTAC-PDA \cite{CPTAC-PDA}, 
HCC-TACE-Seg \cite{HCC-TACE-Seg}, 
TCGA-LIHC \cite{TCGA-LIHC}, 
CMB-CRC \cite{CMB-CRC}, 
TCGA-COAD \cite{TCGA-COAD}, 
TCGA-BLCA \cite{TCGA-BLCA}, 
TCGA-KIRC \cite{TCGA-KIRC}, 
C4KC-KiTS \cite{C4KC-KiTS}, 
CPTAC-CCRCC \cite{CPTAC-CCRCC}, 
TCGA-KIRP \cite{TCGA-KIRP}, 
TCGA-KICH \cite{TCGA-KICH}, 
CMB-LCA \cite{CMB-LCA}, 
CPTAC-LSCC \cite{CPTAC-LSCC}, 
CPTAC-LUAD \cite{CPTAC-LUAD}, 
Lung-PET-CT-Dx \cite{Lung-PET-CT-Dx}, 
TCGA-LUSC \cite{TCGA-LUSC}, 
Anti-PD-1\_Lung \cite{Anti-PD-1_Lung}, 
TCGA-BRCA \cite{TCGA-BRCA}\cite{Breast3}, 
UCSF \cite{ucsf1}\cite{ucsf2}\cite{Breast3}, 
I-SPY 1 \cite{I-SPY1}\cite{Breast3}.\\

\section{Registration details}
Registration was implemented via the 3D Slicer Elastix extension using ITK-Elastix backend.
The rigid and affine stages used Advanced Mattes Mutual Information (32 bins), a 4-level resolution pyramid (shrink factors [8,4,2,1]), and adaptive stochastic gradient descent optimization.
The deformable stage employed a third-order B-spline transform (FinalGridSpacingInVoxels = 10) with 3 resolution levels and up to 2000 iterations per level.
All registrations were visually inspected for anatomical consistency before inclusion.

We performed rigid and affine alignment between CE and NCE scans using the \textbf{3D Slicer} Elastix module, followed by deformable registration for motion-prone organs (e.g., liver, lung). 
All stages used the ITK-elastix backend with the following representative settings: 
\textbf{rigid} (Euler3DTransform) and \textbf{affine} (AffineTransform) stages employed the Advanced Mattes Mutual Information (32 bins) as the similarity metric, a multi-resolution pyramid with 4 levels (shrink factors $[8,4,2,1]$; smoothing sigmas $[3,2,1,0]$ vox), the Adaptive Stochastic Gradient Descent optimizer (max 1024 iterations per level), and linear interpolation for resampling; 
the \textbf{deformable} stage used a third-order B-spline transform (BSpline order $=3$) with 3 resolution levels (shrink factors $[4,2,1]$; smoothing sigmas $[2,1,0]$ vox), max 2000 iterations per level, and a final control-point spacing of 10 voxels (FinalGridSpacingInVoxels$=10$), while final resampling used B-spline interpolation. 
This configuration achieved accurate anatomical correspondence while maintaining computational efficiency, reducing potential supervision bias due to CE/NCE misalignment.

\section{Ethics and Data Availability}
\label{sec:ethical}

We verified the licenses and acquisition procedures of all resources to ensure redistribution compliance.
Dataset access is provided upon submission of an agreement form and approval from the original data sources.
It will be released on our project website under a CC BY-NC 4.0 license, prohibiting private redistribution. All users must obtain permission through the website before downloading.




\end{document}